\documentclass[10pt,twocolumn,letterpaper]{article}
\PassOptionsToPackage{table}{xcolor}
\usepackage[pagenumbers]{cvpr}
\def\confName{CVPR}
\def\confYear{2027}

\usepackage{amsmath,amsfonts,bm}

\def\eqref#1{equation~\ref{#1}}

\def\1{\bm{1}}

\DeclareMathAlphabet{\mathsfit}{\encodingdefault}{\sfdefault}{m}{sl}
\SetMathAlphabet{\mathsfit}{bold}{\encodingdefault}{\sfdefault}{bx}{n}

\usepackage{caption}
\usepackage{enumitem}
\usepackage{booktabs}
\usepackage{multirow}
\usepackage[table]{xcolor}   
\usepackage{graphicx}
\usepackage{lipsum}
\usepackage{duckuments} 
\usepackage{xspace}
\usepackage{makecell}   
\usepackage{pifont}     
\definecolor{cvprblue}{rgb}{0.21,0.49,0.74}
\usepackage[pagebackref,breaklinks,colorlinks,allcolors=cvprblue]{hyperref}
\usepackage{url}
\usepackage{tabularray}
\usepackage{cuted}
\usepackage{tabularx}
\usepackage{array}

\title{SparQ4D: Learning to Act through Sparse 4D Queries}

\author{
Siyuan Xie$^{1}$,
Jinxi Xiao$^{1}$,
Ran Ji$^{1}$,
Yikai Liu$^{1}$,
Mingxi Li$^{1}$, \\
Xuebing Li$^{1}$,
Ye Shi$^{1}$,
Yuxiang Gao$^{2}$,
Jiayuan Gu$^{1}$ \\
$^{1}$ShanghaiTech University \qquad
$^{2}$COCOMatrix \\
{\tt\small shanghaitech.edu.cn \qquad cocomatrix.cn}
}

\begin{document}

\maketitle

\newcommand{\method}{SparQ4D\xspace} 

\begin{strip}
  \centering
  \includegraphics[width=0.82\linewidth]{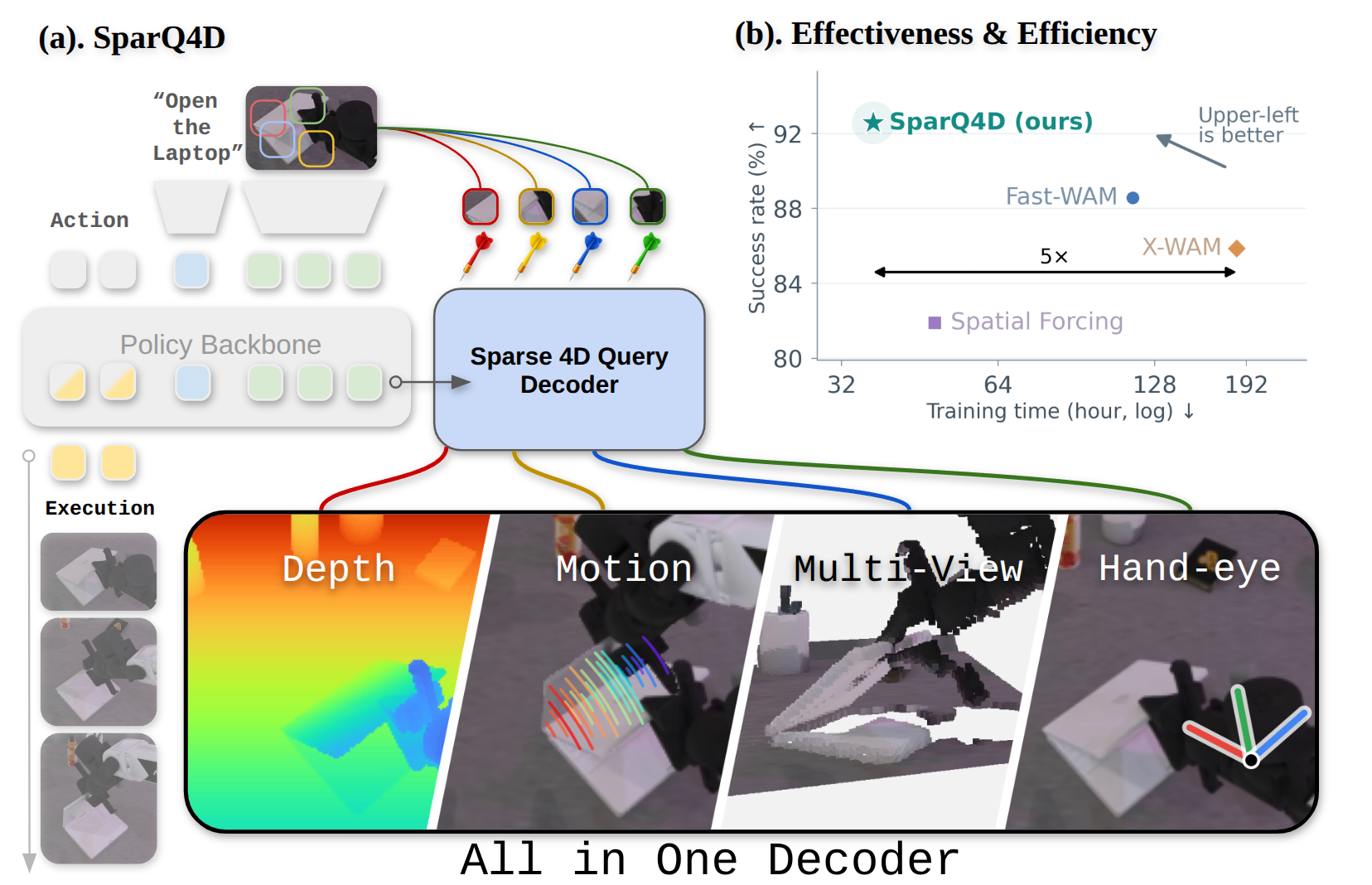}
  \captionsetup{hypcap=false}
  \captionof{figure}{\textbf{Overview of \method.}
  \textbf{(a)} To inject spatial-temporal understanding efficiently, we supervise a policy with sparse 4D queries, decoded by a lightweight decoder attached to its intermediate representations.
  \textbf{(b)} Success rate versus training time on RoboTwin 2.0. \method achieves the highest success rate with fastest training, compared with world action models and representation-alignment methods.}
  \label{fig:teaser}
\end{strip}
\leavevmode

\setlength{\stripsep}{0pt}
\renewcommand{\endabstract}{\par\vspace{12pt}}
\begin{abstract}
    Successful manipulation requires a policy to understand where things are and how they will move, yet action demonstrations leave this spatiotemporal structure implicit. While explicit 4D supervision has proven effective, reducing the cost of dense 4D prediction remains challenging. This paper introduces \method, a lightweight yet expressive decoder that efficiently injects holistic 4D scene understanding into policies and improves success rates. At its core is a query-based decoder attached to a pretrained policy, which casts many spatialtemporal tasks into one unified interface (e.g. depth, 2D/3D point motion, multi-view correspondence, and hand-eye relations). This design enables efficient training and inference without sacrificing expressivity: the policy learns rich 4D structure from a small number of sampled queries during training, while the decoder can be removed at inference with zero additional overhead. Across multiple backbones, simulated environments, and real-world settings, \method achieves state-of-the-art manipulation performance, outperforming competing methods while training up to $5\times$ faster. Probing analyses further reveal that this 4D structure is recoverable from the policy.
\end{abstract}

\section{Introduction}
\label{sec:intro}

Successful manipulation is inherently spatiotemporal: it depends on knowing where things are and how they will move. To hang a mug, a robot must grasp it, align its handle with a rack, and relate wrist and external camera views as it moves. Action demonstrations supervise the desired behavior but leave these relationships implicit. Efficiently teaching a policy this spatiotemporal structure is thus a central question.

One approach to supervising spatiotemporal understanding is through future prediction. World action models (WAMs) jointly model future visual observations and robot actions~\citep{dreamzero}, while X-WAM~\citep{guo2026xwam} extends this paradigm to multi-view RGB-D prediction. Reconstructing the full future supervises scene dynamics richly, but requires dense prediction over many visual tokens, substantially increasing training cost. Fast-WAM~\citep{yuan2026fastwam} removes future generation at inference yet retains strong performance, suggesting that much of the benefit comes from the training signal itself.
Another line injects geometric knowledge into vision-language-action models (VLAs) through pretrained 3D representations: R3DP~\citep{zhang2026r3dp} incorporates foundation-model features via a fast-slow perception system, and Spatial Forcing~\citep{spatialforcing2025} aligns VLA representations with 3D teacher features during training. Although effective, these methods likewise rely on dense, per-token supervision. This leads to our central question: \textbf{can sparse, explicit supervision teach a robot policy both geometry and motion at substantially lower training cost, without additional inference overhead or sacrificing expressivity?}

We introduce \method, a lightweight framework that injects holistic 4D understanding into robot policies through unified sparse supervision (Fig.~\ref{fig:teaser}). Inspired by D4RT~\citep{d4rt}, our key idea is to formulate diverse spatiotemporal tasks as a single query: \emph{where is a selected point at a specified time, in a specified view or coordinate frame?} Given a query specifying the point, time, and reference frame, a lightweight decoder predicts its location. By varying these attributes, the same decoder provides supervision for depth, 2D/3D point motion, multi-view correspondence, and hand-eye relations. We attach this decoder to intermediate VLA representations and jointly optimize auxiliary predictions with action learning, encouraging the policy to encode rich geometric and temporal structure. Crucially, this unified formulation eliminates the need for dense reconstruction. During training, we sample only a small budget of queries across points, times, and views, directly controlling auxiliary decoding cost while retaining diverse spatiotemporal supervision. At inference, the decoder is simply removed, leaving the original policy architecture and computation unchanged. Thus, \method decouples rich training-time supervision from inference-time complexity.

Across simulation benchmarks and real-world tasks, \method achieves state-of-the-art manipulation performance with both $\pi_0$ and $\pi_{0.5}$, reducing failures by up to 60\%. These gains span RoboTwin~2.0~\citep{chen2025robotwin}, LIBERO-Plus~\citep{fei2026liberoplus}, and five real-world tasks, demonstrating the effectiveness of our unified supervision. Beyond performance, \method requires up to $5\times$ fewer training GPU-hours than WAM-based approaches, while introducing no additional inference overhead. Finally, probing analyses reveal recoverable 4D structure, suggesting that the policy itself internalizes this structure rather than relying on the auxiliary decoder.
Our contributions are threefold:
\begin{itemize}
  \item We introduce \method, a lightweight yet expressive framework that injects holistic 4D understanding into robot policies through a unified query-based supervision interface.
  \item We enable efficient training through a small, controllable budget of sparse 4D queries, reducing auxiliary supervision cost without sacrificing expressivity or inference speed.
  \item We demonstrate improved manipulation performance across multiple backbones, simulation environments and the real world. We also systematically examine supervision content, density, placement and internal representation through controlled ablations.
\end{itemize}

\section{Related work}
\label{sec:related_work}


\paragraph{Geometric representations and supervision for policies.}
Prior work incorporates geometric knowledge into robot policies through explicit 3D inputs, such as depth, point clouds, and voxels~\citep{peract,rvt,dp3}, or pretrained 3D features~\citep{zhang2026r3dp}. These methods leverage geometric information at inference time. Alternatively, training-time representation alignment improves spatial understanding without additional geometric inputs at deployment. Spatial Forcing~\citep{spatialforcing2025} aligns VLA representations with pretrained 3D features, while Temporal Forcing~\citep{temporal_forcing} extends it to 4D representations. FLARE~\citep{flare} similarly encourages predictive representations through future latent alignment. Unlike these methods that supervise implicit features, \method directly supervises explicit geometric and motion quantities through a unified, lightweight query decoder.

\paragraph{World modeling and sparse 4D supervision.}
Video-prediction policies such as UniPi and GR-1~\citep{unipi,gr1} use future observations to guide action learning, while world action models (WAMs) jointly model video and actions~\citep{dreamzero,lingbotva}. Recent approaches extend world modeling with 3D/4D information through geometric prediction or alignment~\citep{tesseract,guo2026xwam,4dwam,wam4d}. Fast-WAM~\citep{yuan2026fastwam} removes future generation at inference, while AHA-WAM~\citep{ahawam} accelerates execution through asynchronous world modeling. Despite these advances, future prediction often incurs substantial training cost. Separately, D4RT~\citep{d4rt} introduces unified 4D queries for reconstructing observed video geometry, without forecasting unseen futures. In contrast, \method uses sparse 4D queries to predict geometry and motion from the current observation alone, without video reconstruction or additional inference computation.

\newcommand{\cmark}{\textcolor{green!50!black}{\ding{51}}}
\newcommand{\xmark}{\textcolor{red!70!black}{\ding{55}}}
\newcommand{\pmark}{\textcolor{orange!90!black}{$\circ$}}  

\begin{table}[t]
\centering
\footnotesize
\setlength{\tabcolsep}{2pt}
\renewcommand{\arraystretch}{1.15}
\caption{\textbf{Positioning of \method}, against the methods we compare with. \cmark: satisfied; \pmark: partially satisfied; \xmark: not satisfied.}
\label{tab:positioning}
\begin{tabularx}{\columnwidth}{
    @{}
    >{\raggedright\arraybackslash}p{0.35\columnwidth}
    *{4}{>{\centering\arraybackslash}X}
    @{}
}
\toprule
Method
& Multi-target, one interface
& Explicit target
& No heavy training cost
& No extra inference cost \\
\midrule
X-WAM~\citep{guo2026xwam}
& \pmark & \cmark & \xmark & \xmark \\
Fast-WAM~\citep{yuan2026fastwam}
& \xmark & \cmark & \xmark & \cmark \\
Spatial Forcing~\citep{spatialforcing2025}
& \pmark & \xmark & \xmark & \cmark \\
\midrule
\textbf{\method (ours)}
& \cmark & \cmark & \cmark & \cmark \\
\bottomrule
\end{tabularx}
\end{table}

\section{Method}
\label{sec:method}

\begin{figure*}[t]
  \centering
  \includegraphics[width=\linewidth]{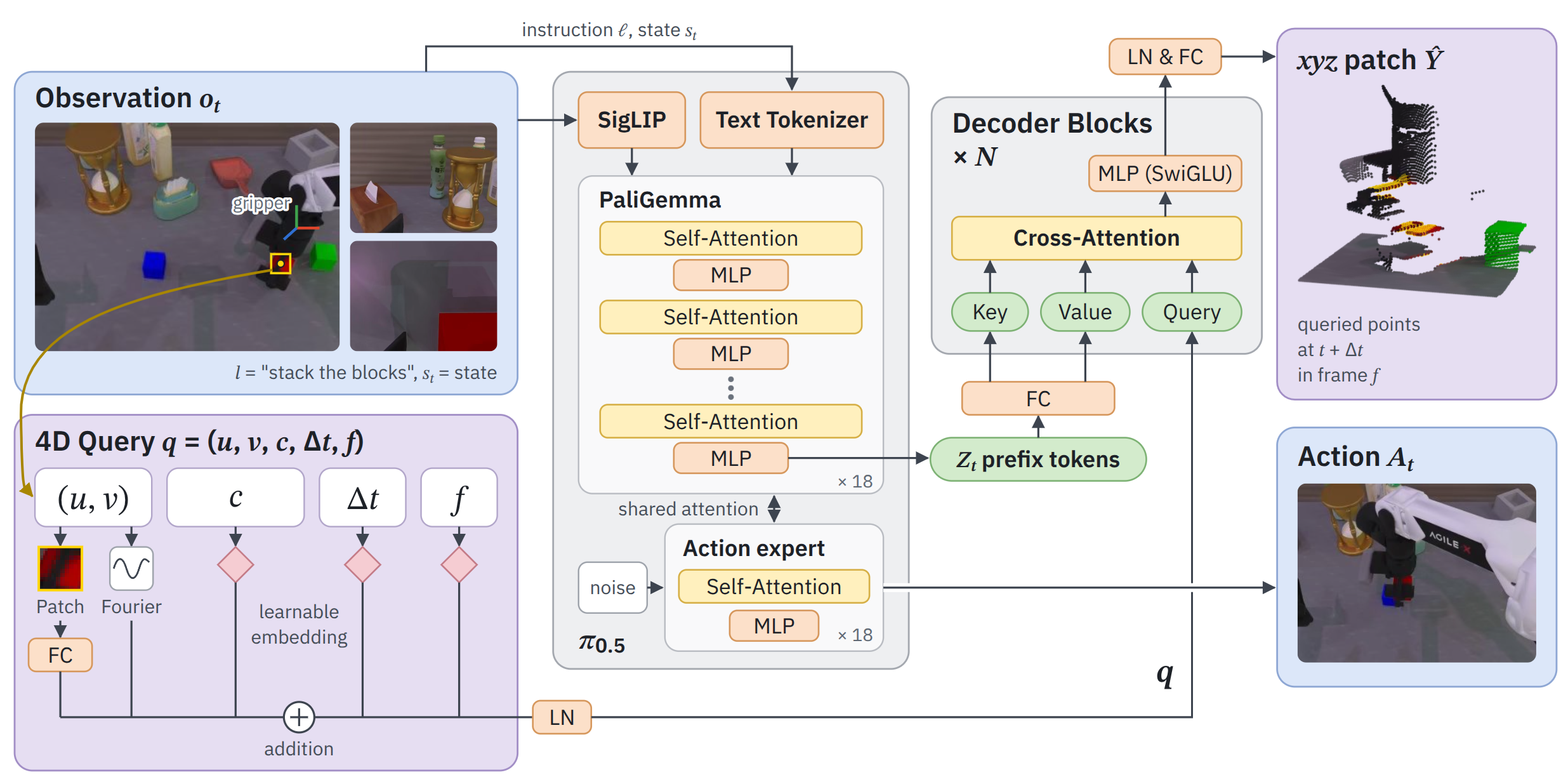}
    \caption{\textbf{Overview of \method (based on $\mathbf{\pi_{0.5}}$).}
    Alongside action prediction, the policy backbone is trained to answer sparse 4D queries $\mathbf{q} = (\mathbf{u}, \mathbf{v}, c, \Delta t, f)$: where the point at pixel $(\mathbf{u}, \mathbf{v})$ of camera $c$ is or will be at time $t + \Delta t$, in frame $f$.
    Each query is embedded as one token, and a lightweight decoder cross-attends to the prefix tokens $\mathbf{Z}_t$ to predict its 3D patch $\hat{\mathbf{Y}}_\mathbf{q}$.
    The decoder is removed after training, leaving the deployed policy unchanged.}
  \label{fig:method}
\end{figure*}

\method adds an auxiliary objective to policy learning: besides predicting actions, its backbone is also trained to answer sparse \emph{4D queries} about the current observation (Fig.~\ref{fig:method}). 
Designing this objective involves two main questions: what to ask the policy and how to read out its answers.
Sec.~\ref{sec:method:queries} answers the first by defining the query that unifies depth, motion, multi-view, and hand-eye supervision. Sec.~\ref{sec:method:decoder} answers the second with a lightweight decoder on the policy's backbone. Sec.~\ref{sec:method:training} presents details in training and inference.

\paragraph{Background.}
We build on flow-matching-based vision-language-action (VLA) policies, exemplified by the $\pi$ series~\citep{black2026pi0visionlanguageactionflowmodel,pi05}.
Given the proprioceptive state $\mathbf{s}_t$, camera images $\mathbf{I}_t = \{\mathbf{I}^c_t\}_{c \in \mathcal{C}}$, and a language instruction $\ell$, a policy $\pi_\theta$ predicts an action chunk $\mathbf{A}_t = [\mathbf{a}_t, \ldots, \mathbf{a}_{t+H-1}]$ using flow matching~\citep{lipman2023flowmatchinggenerativemodeling}:
\begin{equation}
\mathcal{L}_{\text{act}} = \mathbb{E}_{\tau, \boldsymbol{\epsilon}} \big\| \mathbf{v}_\theta\big(\tau \mathbf{A}_t + (1-\tau)\boldsymbol{\epsilon};\, \mathbf{s}_t, \mathbf{I}_t, \ell\big) - (\mathbf{A}_t - \boldsymbol{\epsilon}) \big\|^2,
\label{eq:action_loss}
\end{equation}
where $\boldsymbol{\epsilon}$ is standard Gaussian noise and $\tau \in [0, 1]$. In our bimanual setting, $\mathcal{C}$ consists of a head camera and one wrist camera per arm, and $\mathcal{G}$ denotes the two grippers. The policy employs a vision-language backbone that produces intermediate representations $\mathbf{Z}_t$, coupled with an action expert for action generation. \method leaves $\mathcal{L}_{\text{act}}$ unchanged and introduces an auxiliary objective on $\mathbf{Z}_t$.

\subsection{4D queries}
\label{sec:method:queries}

\paragraph{Query.}
We define a 4D query at time $t$ as
\begin{equation}
\mathbf{q} = (\mathbf{u}, \mathbf{v}, c, \Delta t, f),
\label{eq:query}
\end{equation}
which asks: \emph{where \textbf{is or will} the point currently seen at pixel $(\mathbf{u}, \mathbf{v})$ of camera $c$ be at time $t + \Delta t$, in reference frame $f$?}
Here $c \in \mathcal{C}$ is the source camera, $\Delta t \in \{0, \ldots, H{-}1\}$ is the time offset, and $f \in \mathcal{F} = \mathcal{C} \cup \mathcal{G}$ is the reference camera or gripper frame at time $t$.
The target $\mathbf{Y}_{\mathbf{q}}$ is the queried points' 3D patch at $t + \Delta t$, expressed in frame $f$.
Unlike D4RT~\citep{d4rt}, which reconstructs points from observed video frames, our formulation requires predicting future motion from the current observation alone, without access to future frames or camera poses.

\begin{table}[t]
\centering
\footnotesize
\setlength{\tabcolsep}{3pt}
\renewcommand{\arraystretch}{1.15}
\caption{\textbf{Query types.} Each type is an instance of
$\mathbf{q} = (\mathbf{u}, \mathbf{v}, c, \Delta t, f)$.
Hand--eye queries use the head camera as source, since
wrist cameras are rigidly mounted on their grippers;
the other types use any camera.}
\label{tab:query_types}
\begin{tabularx}{\columnwidth}{
    @{}
    l
    c
    >{\centering\arraybackslash}X
    l
    @{}
}
\toprule
Type & Offset $\Delta t$ & Reference $f$ & Ability \\
\midrule
Depth
& $0$ & $c$ & metric structure \\
Motion
& $1,\ldots,H{-}1$ & $c$ & future motion \\
Multi-view
& $0$ & $c' \in \mathcal{C}\setminus\{c\}$ & relative camera pose \\
Hand-eye
& $0$ & $g \in \mathcal{G}$ & gripper pose \\
\bottomrule
\end{tabularx}
\end{table}

\paragraph{One interface for all queries.}
The same question, with different $\Delta t$ and $f$, covers the four kinds of geometry that manipulation relies on (Tab.~\ref{tab:query_types}):
\begin{itemize}[leftmargin=*, topsep=2pt, itemsep=1pt, parsep=0pt]
\item \textbf{Depth}: where is this point \emph{now}, in the current camera frame?
\item \textbf{Motion}: where \emph{will} this point be at $t + \Delta t$, in the current camera frame?
\item \textbf{Multi-view}: where is this point now, in \emph{another} camera's frame?
\item \textbf{Hand-eye}: where is this point now, in a \emph{gripper}'s frame?
\end{itemize}

\subsection{Query decoder}
\label{sec:method:decoder}

\paragraph{Architecture.}
To answer 4D queries, we introduce a lightweight transformer decoder $\mathcal{D}_\phi$ that reads from the policy's intermediate representations $\mathbf{Z}_t$.
Each query is encoded as a single token by summing a linear embedding of the $p \times p$ RGB patch around $(\mathbf{u}, \mathbf{v})$ in $\mathbf{I}^c_t$, a positional encoding of $(\mathbf{u}, \mathbf{v})$, and learned embeddings of $c$, $\Delta t$, and $f$.
Query tokens cross-attend to the prefix tokens $\mathbf{Z}_t$ (image tokens from all cameras and language tokens), but not to each other, allowing independent decoding.
A linear layer maps each decoded token to a predicted patch $\hat{\mathbf{Y}}_\mathbf{q} = \mathcal{D}_\phi(\mathbf{q}; \mathbf{Z}_t)$.
The decoder adds only 1.6\% to the policy's parameter count.

\subsection{Training and inference}
\label{sec:method:training}

During training, we jointly optimize action learning and sparse 4D supervision. At each time step, we sample $M$ queries, distributed equally among the four types. Spatial locations $(\mathbf{u}, \mathbf{v})$ are sampled preferentially near depth discontinuities, with uniform sampling to ensure broader coverage. For motion queries, $\Delta t$ is sampled uniformly from valid future offsets (App.A). 

We optimize the policy and decoder with
\begin{align}
\mathcal{L}_{\text{geo}} &= \frac{1}{|\mathcal{Q}|} \sum_{\mathbf{q} \in \mathcal{Q}} \frac{1}{3p^2} \big\| \hat{\mathbf{Y}}_\mathbf{q} - \mathbf{Y}_\mathbf{q} \big\|_1, \label{eq:geo_loss} \\
\mathcal{L} &= \mathcal{L}_{\text{act}} + \lambda\, \mathcal{L}_{\text{geo}}, \label{eq:total_loss}
\end{align}
where $\mathcal{Q}$ contains valid queries, excluding those with invalid depth or unavailable correspondences. The geometry loss averages the absolute error over all $3p^2$ coordinates of each predicted patch, with $\lambda$ controlling the strength of auxiliary supervision. Both objectives update the policy backbone, encouraging it to learn geometric and temporal structure alongside action prediction.

At inference, the decoder is discarded, leaving the original VLA inputs, architecture, and computational cost unchanged. Since \method operates on intermediate vision-language representations, it can be applied to different backbones without modifying their action prediction mechanisms.

\section{Experiments}
\label{headings}

\begin{figure*}[t]
\centering
\setlength{\tabcolsep}{1pt}
\renewcommand{\arraystretch}{1.0}
\newcommand{\imgw}{\dimexpr(\linewidth-8\tabcolsep)/5\relax}

\newcommand{\imgcell}[2]{%
  \parbox[t]{\imgw}{%
    \centering
    \includegraphics[width=\linewidth]{#1} \\[-1pt]
    {\scriptsize #2\par}%
  }%
}

\begin{tabular}{@{}c c c c c@{}}

\imgcell{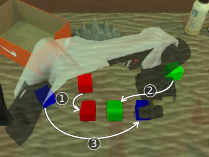}{Blocks Ranking RGB} &
\imgcell{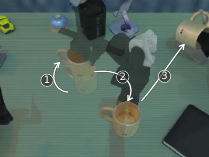}{Hanging Mug} &
\imgcell{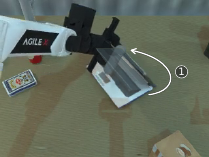}{Open Laptop} &
\imgcell{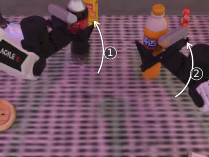}{Pick Diverse Bottles} &
\imgcell{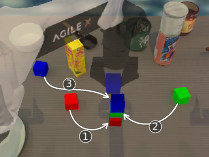}{Stack Blocks Three} \\[6pt]

\imgcell{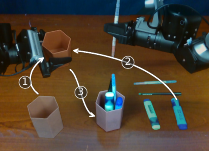}{Fill Pen Holder} &
\imgcell{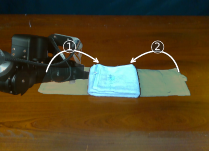}{Fold Towel} &
\imgcell{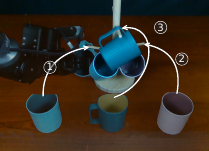}{Hang Mugs} &
\imgcell{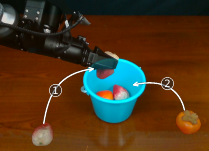}{Place Fruits} &
\imgcell{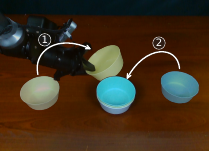}{Stack Bowls} \\[6pt]

\imgcell{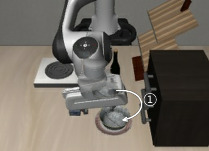}{Place Bowl} &
\imgcell{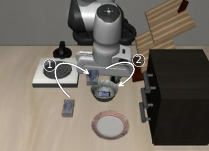}{Place Cream Cheese} &
\imgcell{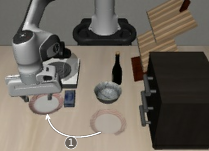}{Push Plate} &
\imgcell{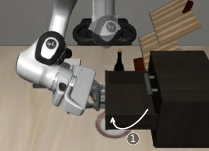}{Open Drawer} &
\imgcell{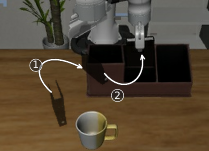}{Stow Book} \\

\end{tabular}

\caption{Representative task setups in
RoboTwin~2.0 (top row),
the real world (middle row),
and LIBERO-Plus (bottom row).}
\label{fig:qualitative_tasks}
\end{figure*}

Our experiments test both effectiveness and efficiency, and examine why \method works. We organize them around five questions:
\begin{itemize}
    \item \textbf{RQ1 (Effectiveness, Simulation):} Does \method improve manipulation success over strong VLA and WAM baselines in simulation?
    \item \textbf{RQ2 (Effectiveness, Real World):} Does this improvement hold on a physical robot?
    \item \textbf{RQ3 (Efficiency):} Is \method cheaper to train and faster to deploy than dense video-based and implicit-alignment baselines?
    \item \textbf{RQ4 (Ablation):} Which query types and architectural choices drive this improvement?
    \item \textbf{RQ5 (Analysis):} Does \method make the VLA representation geometry-aware? 
\end{itemize}

\subsection{Experimental setup}
\label{sec:setup}

\paragraph{Training.}
On RoboTwin~2.0~\citep{chen2025robotwin}, we use 50 clean-setting and 500 randomized-setting demonstrations per task, training for 5 epochs on eight NVIDIA H20 GPUs, following Fast-WAM~\citep{yuan2026fastwam}. On LIBERO-Plus\cite{fei2026liberoplus}, we train on 130 tasks with 50 demonstrations each. In simulation, depth queries are supervised with rendered ground-truth depth. For real-world evaluation, we collect 50 demonstrations for each of five tasks and supervise depth queries with MoGe-3~\citep{kong2026finedetailmonoculargeometryestimation} estimates (see App.C for details). Representative tasks are shown in Fig.~\ref{fig:qualitative_tasks}.

\paragraph{Evaluation.} In simulation, we roll out each method 100 times per task under both \textit{Clean} and \textit{Randomized} settings. On the real robot, we roll out 25 episodes per task, due to the time-intensive nature of real-world evaluation. We report success rate (SR) throughout. For real-world tasks, we additionally report progress score (PS), a partial-credit metric defined in App.C.

\paragraph{Baselines.} All baselines are initialized from their publicly released
pretrained weights and fine-tuned on RoboTwin under the same protocol as ours.
We compare against three groups: (i) $\pi_0$ and $\pi_{0.5}$~\citep{black2026pi0visionlanguageactionflowmodel,pi05},
action-only VLA baselines; (ii) Spatial Forcing~\citep{spatialforcing2025}, an
implicit-alignment baseline (Sec.~\ref{sec:related_work}) built on $\pi_0$, for
which we use the implementation from XPolicyLab~\citep{community2026xpolicylab};
and (iii) X-WAM~\citep{guo2026xwam} and Fast-WAM~\citep{yuan2026fastwam}, two WAM baselines
built on the Wan2.2~\citep{wan2025} video foundation model.


\subsection{RQ1: Does \method improve manipulation success in simulation?}
\label{sec:rq1}

Tables~\ref{tab:main_results} and~\ref{tab:libero_plus} report success rates on RoboTwin~2.0 and LIBERO-Plus, respectively.

\paragraph{Improvements across benchmarks and backbones.}
\method improves both $\pi_0$ and $\pi_{0.5}$ across the two simulation benchmarks. On RoboTwin~2.0, it increases average success rates by roughly 25 and 15 percentage points for the two backbones, respectively, under both Clean and Randomized settings. Notably, \method with $\pi_{0.5}$ outperforms strong WAM baselines, achieving the highest average success rate in both settings without future video generation. On LIBERO-Plus, \method improves $\pi_0$ by 19.3 points, surpassing Spatial Forcing, while the stronger $\pi_{0.5}$ version achieves the highest aggregate success rate of 85.0\%, with a smaller gain over its already competitive baseline (84.4\%). These results demonstrate the effectiveness of sparse 4D supervision across distinct simulation benchmarks and VLA backbones, although the magnitude of improvement varies.

\paragraph{Gains are pronounced in geometry-sensitive settings.}
On RoboTwin~2.0, \method brings particularly large improvements to tasks requiring precise spatial alignment between the gripper, object, and target. With $\pi_{0.5}$, hanging-mug improves from 18\% to 70\% in Clean and from 17\% to 57\% in Randomized settings, while pick-diverse-bottles gains 23 points under randomization. On LIBERO-Plus, a similar benefit emerges under spatial perturbations: with $\pi_0$, success rates under camera and robot perturbations improve by 50.6 and 45.7 points, respectively. Even with the stronger $\pi_{0.5}$, robot perturbations improve by 8.9 points, although gains across other perturbation categories are mixed. Together, these results suggest that explicit 4D supervision is particularly beneficial for manipulation requiring accurate spatial reasoning and alignment.

\providecommand{\srpair}[2]{#1\,/\,#2}

\begin{table*}[t]
\centering
\small
\caption{Results on 7 RoboTwin tasks. Each per-task cell reports
the number of successful trials out of 100 (Clean / Random).
Avg columns report the mean success rate (\%) over the 7 tasks.
\textbf{Bold} and \textit{italics} denote the best and second-best
result in each column, respectively (ties are all marked).
Rows shaded in gray are ours. All models are trained for 5 epochs.
We further evaluate on the full 50-task benchmark and observe
consistent improvements (Appendix B).} 
\label{tab:main_results}

\setlength{\tabcolsep}{2pt}
\renewcommand{\arraystretch}{1.15}

\begin{tabularx}{\textwidth}{
    @{}
    >{\raggedright\arraybackslash}p{0.165\textwidth}
    >{\raggedright\arraybackslash}p{0.145\textwidth}
    *{2}{>{\centering\arraybackslash}p{0.075\textwidth}}
    *{7}{>{\centering\arraybackslash}X}
    @{}
}
\toprule
& &
\multicolumn{2}{c}{\textbf{Avg.\ SR $\uparrow$}}
& \multicolumn{7}{c}{
    \textbf{7 tasks} (Clean / Rand SR) $\uparrow$
} \\
\cmidrule(lr){3-4}
\cmidrule(l){5-11}

\textbf{Method}
& \textbf{Backbone (Type)}
& Clean & Rand
& \shortstack{block\\ranking\\rgb}
& \shortstack{hanging\\mug}
& \shortstack{open\\laptop}
& \shortstack{pick\\diverse\\bottles}
& \shortstack{rotate\\qrcode}
& \shortstack{stack\\blocks\\three}
& \shortstack{place\\shoe} \\
\midrule

$\pi_0$
& $\pi_0$ (VLA)
& 60.71 & 57.86
& \srpair{70}{73}
& \srpair{14}{11}
& \srpair{76}{76}
& \srpair{49}{31}
& \srpair{63}{63}
& \srpair{77}{75}
& \srpair{76}{76} \\

Spatial Forcing
& $\pi_0$ (VLA)
& 82.00 & 84.43
& \srpair{93}{87}
& \srpair{\textbf{80}}{\textbf{77}}
& \srpair{67}{70}
& \srpair{47}{73}
& \srpair{\textbf{100}}{\textbf{97}}
& \srpair{87}{87}
& \srpair{\textbf{100}}{\textbf{100}} \\

\rowcolor{gray!15}
{\method} (ours)
& $\pi_0$ (VLA)
& 85.43 & 85.29
& \srpair{96}{90}
& \srpair{63}{56}
& \srpair{93}{\textbf{100}}
& \srpair{70}{79}
& \srpair{90}{86}
& \srpair{90}{88}
& \srpair{96}{98} \\

\midrule

$\pi_{0.5}$
& $\pi_{0.5}$ (VLA)
& 77.57 & 75.00
& \srpair{82}{85}
& \srpair{18}{17}
& \srpair{90}{96}
& \srpair{81}{71}
& \srpair{89}{87}
& \srpair{91}{76}
& \srpair{92}{93} \\

\rowcolor{gray!15}
{\method} (ours)
& $\pi_{0.5}$ (VLA)
& \textbf{92.57} & \textbf{90.14}
& \srpair{\textbf{100}}{\textbf{97}}
& \srpair{\textit{70}}{57}
& \srpair{\textit{96}}{96}
& \srpair{\textit{86}}{\textbf{94}}
& \srpair{\textbf{100}}{\textit{95}}
& \srpair{\textbf{96}}{\textit{92}}
& \srpair{\textbf{100}}{\textbf{100}} \\

\midrule

X-WAM
& Wan2.2 (WAM)
& 85.86 & 86.14
& \srpair{\textit{99}}{\textit{95}}
& \srpair{46}{55}
& \srpair{\textit{96}}{\textit{97}}
& \srpair{\textbf{91}}{\textit{92}}
& \srpair{84}{83}
& \srpair{88}{82}
& \srpair{\textit{97}}{\textit{99}} \\

Fast-WAM
& Wan2.2 (WAM)
& \textit{88.57} & \textit{89.43}
& \srpair{\textbf{100}}{\textbf{97}}
& \srpair{58}{\textit{62}}
& \srpair{\textbf{98}}{\textit{97}}
& \srpair{80}{85}
& \srpair{\textit{93}}{89}
& \srpair{\textit{95}}{\textbf{97}}
& \srpair{96}{\textit{99}} \\

\bottomrule
\end{tabularx}
\end{table*}



\begin{table}[t]
\centering
\caption{\textbf{Results on LIBERO-Plus.}
Success rates (\%) under seven perturbation
categories. Cam., Rob., Lang., and BG denote
camera, robot, language, and background,
respectively. Avg. denotes the aggregate
success rate across all perturbations.
\textbf{Bold} and \textit{italics} indicate
the best and second-best results in each column.
Gray rows indicate our method.}
\label{tab:libero_plus}

\scriptsize
\setlength{\tabcolsep}{1pt}
\renewcommand{\arraystretch}{1.1}

\begin{tabularx}{\columnwidth}{
    @{}
    >{\raggedright\arraybackslash}p{0.25\columnwidth}
    *{8}{>{\centering\arraybackslash}X}
    @{}
}
\toprule
\textbf{Method}
& \textbf{Avg. $\uparrow$} & Cam. & Rob. & Lang.
& Light & BG & Noise & Layout \\
\midrule

$\pi_0$
& 53.6 & 13.8 & 6.0 & 58.8
& 85.0 & 81.4 & 79.0 & 68.9 \\

Spatial Forcing
& 66.0 & \textit{74.2} & 32.5 & 57.3
& 85.8 & 82.3 & 72.5 & 67.1 \\

\rowcolor{gray!15}
\method ($\pi_0$)
& 72.9 & 64.4 & 51.7 & 66.6
& 91.7 & 88.2 & \textit{82.0} & 75.2 \\

\midrule

$\pi_{0.5}$
& \textit{84.4} & \textbf{78.4} & \textit{73.6} & 80.8
& \textit{96.2} & \textit{94.1} & \textbf{89.0} & \textit{84.5} \\

\rowcolor{gray!15}
\method ($\pi_{0.5}$)
& \textbf{85.0} & 73.8 & \textbf{82.5} & 85.2
& \textbf{96.4} & \textbf{96.8} & 81.2 & \textbf{85.7} \\

\midrule

X-WAM
& 73.5 & 61.0 & 61.9 & \textit{86.8}
& 94.3 & 92.9 & 49.9 & 80.6 \\

Fast-WAM
& 62.2 & 27.9 & 60.9 & \textbf{89.3}
& 86.7 & 49.3 & 51.1 & 74.8 \\

\bottomrule
\end{tabularx}
\end{table}


\subsection{RQ2: Does \method improve manipulation success in the real world?}
\label{sec:rq2}

The improvements observed in simulation also extend to challenging real-world manipulation. We evaluate five tasks under diverse tabletop layouts, varying object positions, orientations, and spatial arrangements. These tasks require precise, often multi-step manipulation, such as inserting four small pens into a holder or hanging three mugs with different orientations. All methods are evaluated on the same 25 configurations per task to ensure fair comparison (App.C). Table~\ref{tab:real_world} reports success rate (SR) and progress score (PS).

\paragraph{Consistent improvements in real-world manipulation.}
\method improves both backbones in average SR and PS, whereas Spatial Forcing provides little improvement over $\pi_0$. With $\pi_{0.5}$, \method achieves the best overall performance, raising average SR from 60\% to 71\% while improving or matching the backbone on every task. We also observe that X-WAM and Fast-WAM perform well on object layouts resembling their training demonstrations, but their success rates drop sharply under even modest changes in object placement, suggesting weaker robustness to spatial variations. These results highlight the value of explicit 4D supervision for robust manipulation under diverse real-world configurations.

\providecommand{\tn}[1]{%
    \begin{tabular}[b]{@{}c@{}}#1\end{tabular}}

\begin{table*}[t]
\centering
\small
\caption{Real-world results. Avg.\ columns report
success rate (\%) and progress score averaged
over all tasks. Each task cell reports
success rate / progress score. Higher is better.
\textbf{Bold}: best in each backbone group.
Rows shaded in gray are ours.
All models are trained with the same number of seen samples.}
\label{tab:real_world}

\setlength{\tabcolsep}{3pt}
\renewcommand{\arraystretch}{1.05}

\begin{tabularx}{\textwidth}{
    @{}
    >{\raggedright\arraybackslash}p{0.13\textwidth}
    *{2}{>{\centering\arraybackslash}p{0.105\textwidth}}
    *{5}{>{\centering\arraybackslash}X}
    @{}
}
\toprule
&
\multicolumn{2}{c}{\textbf{Avg.}}
& \multicolumn{5}{c}{\textbf{Tasks} (Success Rate / Progress Score)} \\
\cmidrule(lr){2-3}
\cmidrule(l){4-8}

\textbf{Method}
& \shortstack{\textbf{Success}\\\textbf{Rate} $\uparrow$}
& \shortstack{\textbf{Progress}\\\textbf{Score} $\uparrow$}
& \tn{fill\\pen-holder}
& \tn{fold\\towel}
& \tn{hang\\mugs}
& \tn{place\\fruits}
& \tn{stack\\bowls} \\
\midrule

$\pi_0$
& 34.4 & 43.4
& \textbf{12.0} / 23.6
& 44.0 / 58.0
& 16.0 / 26.8
& 48.0 / 54.0
& 52.0 / 54.4 \\

Spatial Forcing
& 31.2 & 37.6
& 8.0 / 14.8
& 28.0 / 34.0
& 20.0 / 33.2
& 56.0 / 62.0
& 44.0 / 44.0 \\

\rowcolor{gray!15}
\method ($\pi_0$)
& \textbf{44.8}
& \textbf{56.0}
& \textbf{12.0} / \textbf{27.8}
& \textbf{48.0} / \textbf{60.0}
& \textbf{28.0} / \textbf{46.4}
& \textbf{64.0} / \textbf{73.6}
& \textbf{72.0} / \textbf{72.0} \\

\midrule

$\pi_{0.5}$
& 60.8 & 70.5
& 52.0 / 59.6
& 44.0 / 62.0
& 32.0 / 50.0
& 84.0 / 88.8
& \textbf{92.0} / \textbf{92.0} \\

\rowcolor{gray!15}
\method ($\pi_{0.5}$)
& \textbf{71.2}
& \textbf{79.6}
& \textbf{56.0} / \textbf{67.6}
& \textbf{60.0} / \textbf{76.0}
& \textbf{52.0} / \textbf{65.2}
& \textbf{96.0} / \textbf{97.2}
& \textbf{92.0} / \textbf{92.0} \\

\midrule

X-WAM
& 22.4 & 29.2
& 0.0 / 0.0
& 8.0 / 10.0
& 12.0 / 30.8
& 68.0 / 76.4
& 24.0 / 28.8 \\

Fast-WAM
& 30.4 & 38.2
& 4.0 / 19.4
& 20.0 / 24.0
& 28.0 / 39.2
& 40.0 / 47.2
& 60.0 / 61.2 \\

\bottomrule
\end{tabularx}
\end{table*}


\subsection{RQ3: Is \method more efficient to train and deploy?}
\label{sec:rq3}

\paragraph{\method is cheaper to train and deploy.}
Compared with the two WAMs, \method ($\pi_{0.5}$) trains 3--5$\times$ faster and runs inference 1.7--6.6$\times$ faster, with about a quarter of their parameters and 24--41\% of their training memory (Tab.~\ref{tab:cost}). Importantly, these efficiency gains do not come at the expense of performance: with sparse 4D supervision, \method outperforms both WAM counterparts on RoboTwin~2.0 and LIBERO-Plus, demonstrating the effectiveness of explicit geometric and motion supervision as an efficient alternative to dense future prediction. Since the query decoder is used only during training, it adds no inference overhead to the backbone. Compared with Spatial Forcing, whose costs are otherwise similar, \method also avoids a frozen VGGT teacher~\citep{wang2025vggt}, reducing training overhead and implementation complexity.

\begin{table}[t]
\centering
\footnotesize
\setlength{\tabcolsep}{2pt}
\renewcommand{\arraystretch}{1.15}
\caption{\textbf{Training and inference cost.}
All measurements use the real-world setting (10k training steps,
batch size 64). Params include trainable and frozen parameters.
Training VRAM is in GB, training time is measured on eight
NVIDIA H20 GPUs, and inference latency on a single RTX 3090.
Inference latency is reported as mean $\pm$ std over five runs.
Both WAMs use gradient accumulation over two microbatches of 32,
as a full batch of 64 exceeds the available GPU memory for X-WAM.
Gray rows indicate our method.}
\label{tab:cost}
\begin{tabularx}{\columnwidth}{
    @{}
    >{\raggedright\arraybackslash}X
    cccc
    @{}
}
\toprule
\textbf{Method}
& \textbf{Params}
& \textbf{VRAM}
& \shortstack{\textbf{Training}\\\textbf{time (h)} $\downarrow$}
& \shortstack{\textbf{Inference}\\\textbf{latency (ms)} $\downarrow$} \\
\midrule
$\pi_0$
& 3.24B & 88 & 31 & $127.14 \pm 0.73$ \\

Spatial Forcing ($\pi_0$)
& 4.15B & 144 & 54 & $127.68 \pm 0.88$ \\

\rowcolor{gray!15}
\method ($\pi_0$)
& 3.29B & 92 & 31 & $127.36 \pm 0.75$ \\

\midrule
$\pi_{0.5}$
& 3.35B & 95 & 32 & $145.98 \pm 0.65$ \\

\rowcolor{gray!15}
\method ($\pi_{0.5}$)
& 3.40B & 99 & 32 & $145.10 \pm 0.77$ \\

\midrule
Fast-WAM (Wan2.2)
& 12.40B & 399 & 101 & $249.25 \pm 2.59$ \\

X-WAM (Wan2.2)
& 13.05B & 679 & 160 & $960.71 \pm 5.64$ \\
\bottomrule
\end{tabularx}
\end{table}

\begin{table}[t]
\centering
\footnotesize
\setlength{\tabcolsep}{3pt}
\renewcommand{\arraystretch}{1.15}
\caption{\textbf{Ablation on RoboTwin (SR, \%).}
All variants use $\pi_{0.5}$ and are trained
for 5 epochs on 7 tasks, as in
Table~\ref{tab:main_results}.
We ablate query type (green), query count (blue),
and cross-attention layer (red).
The default is 96 queries at layer 18,
with all query types enabled.
DPT uses multi-layer features from layers 4, 8, 12, and 18.
Colored cells indicate the varied factor.
\textbf{Bold}: best.}
\label{tab:ablation}
\begin{tabularx}{\columnwidth}{
    @{}
    >{\raggedright\arraybackslash}p{0.4\columnwidth}
    *{4}{>{\centering\arraybackslash}X}
    @{}
}
\toprule
\textbf{Configuration}
& \textbf{\#Q}
& \textbf{Layer}
& \textbf{Clean}
& \textbf{Rand} \\
\midrule
Baseline
& -- & -- & 77.57 & 75.00 \\
DPT Head
& -- & 4/8/12/18 & 85.85 & 81.43 \\
\midrule

\cellcolor{green!15}Depth only
& 96 & 18 & 88.57 & 87.57 \\
\cellcolor{green!15}Motion only
& 96 & 18 & 89.43 & 83.71 \\
\cellcolor{green!15}Hand-eye only
& 96 & 18 & 87.57 & 86.00 \\
\cellcolor{green!15}Multi-view only
& 96 & 18 & 88.00 & 86.57 \\
\cellcolor{green!15}\textbf{Full}
& 96 & 18 & \textbf{92.57} & \textbf{90.14} \\
\midrule

Full
& \cellcolor{blue!12}24
& 18 & 84.86 & 85.43 \\
Full
& \cellcolor{blue!12}48
& 18 & 89.14 & 88.14 \\
\textbf{Full}
& \cellcolor{blue!12}96
& 18 & \textbf{92.57} & \textbf{90.14} \\
Full
& \cellcolor{blue!12}192
& 18 & 91.43 & 89.00 \\
\midrule

Full
& 96 & \cellcolor{red!12}6
& 82.29 & 79.14 \\
Full
& 96 & \cellcolor{red!12}12
& 87.14 & 88.00 \\
\textbf{Full}
& 96 & \cellcolor{red!12}18
& \textbf{92.57} & \textbf{90.14} \\
\bottomrule
\end{tabularx}
\end{table}

\subsection{RQ4: Which design choices drive this improvement?}
\label{sec:rq4}

Table~\ref{tab:ablation} examines the effects of query types, query budget, and decoder attachment layer.

\paragraph{Sparse, unified supervision works best.}
As a baseline, we optimized a dense DPT prediction head with the policy backbone, which improves over the vanilla $\pi_{0.5}$ but underperforms even depth-only sparse queries, suggesting that dense prediction is not necessary for effective geometric supervision. Early experiments also showed
that uniform query sampling was less effective than this targeted strategy, highlighting the importance of where queries are sampled. From the supervision perspective, each query type independently improves success by 9-13 points, while combining all four achieves the best results (92.57\% Clean, 90.14\% Rand), indicating complementary benefits. Motion-only supervision performs relatively worse under randomization, potentially due to the uncertainty of future point locations.

\paragraph{A small query budget suffices.}
Increasing the query count from 24 to 96 steadily improves success, while 192 queries bring no further gains. This suggests that sparse supervision with a modest query budget is sufficient, avoiding the cost of dense prediction.

\paragraph{Deeper supervision is more effective.}
Performance improves monotonically as the decoder is attached to deeper layers, peaking at layer 18. Deeper attachment allows auxiliary gradients to shape more backbone layers. Sec.~\ref{sec:rq5} further examines the learned representations.



\newcolumntype{L}{>{\raggedright\arraybackslash\scriptsize\bfseries}m{0.12\linewidth}}
\newcolumntype{C}{>{\centering\arraybackslash\scriptsize}m{0.20\linewidth}}
\providecommand{\ph}[3]{\fcolorbox{black!40}{black!5}{\parbox[c][#2][c]{\dimexpr#1-2\fboxsep-2\fboxrule\relax}{\centering\scriptsize #3}}}

\begin{figure*}[!t]
\centering
\setlength{\tabcolsep}{1pt}
\renewcommand{\arraystretch}{1.0}
\newcommand{\probedir}{figs/experiments/probings}   
\newcommand{\pw}{0.205\linewidth}      

\newlength{\probeh}
\settoheight{\probeh}{\includegraphics[width=\pw]{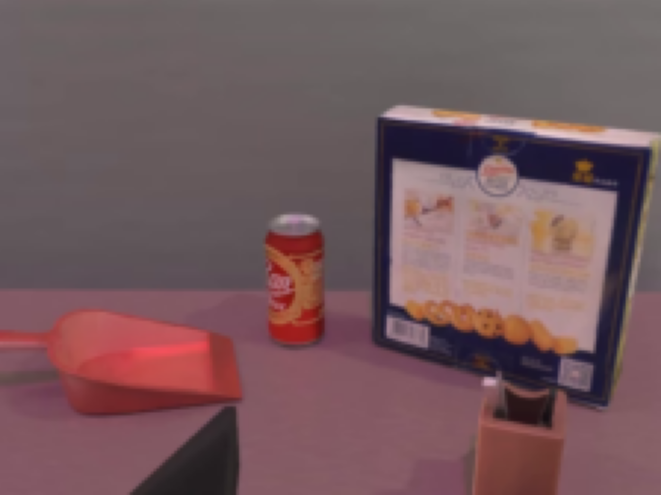}}

\newcommand{\pimg}[1]{\includegraphics[width=\pw]{\probedir/#1.png}}
\newcommand{\plabel}[1]{\raisebox{\dimexpr 0.5\probeh - 0.5ex\relax}{\small\textbf{#1}}}
\newcommand{\phead}[1]{\small #1}

\begin{tabular}{@{}r c c c c@{}}
\plabel{Input View} &
\pimg{input_depth} & \pimg{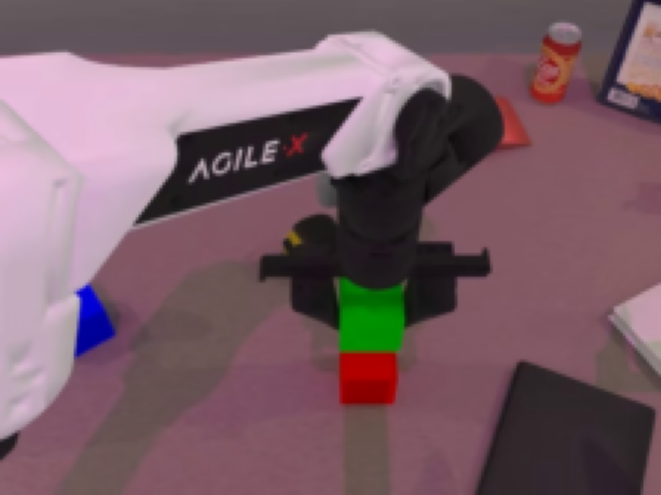} & \pimg{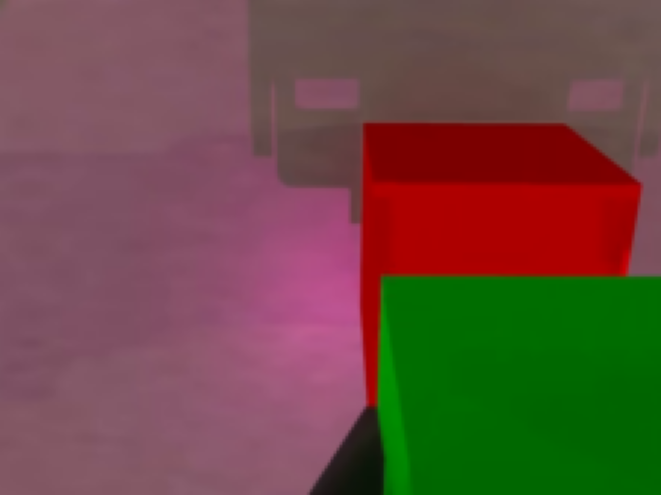} & \pimg{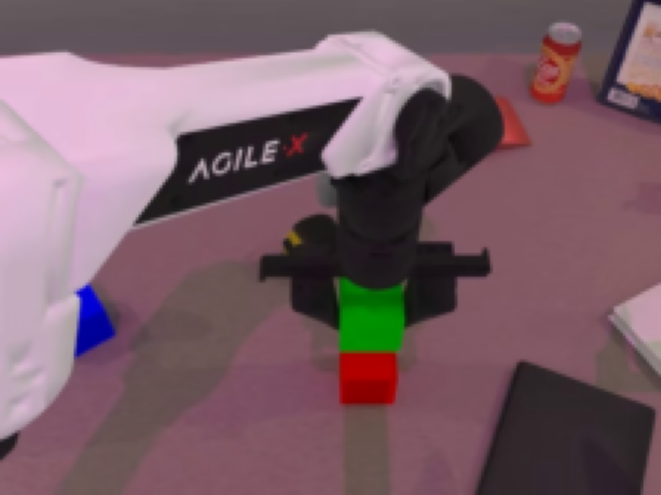} \\[1pt]
{\small\textbf{Query Type}} &
\phead{depth} & \phead{motion} & \phead{multi-view} & \phead{hand-eye} \\[2pt]
\plabel{Ours} &
\pimg{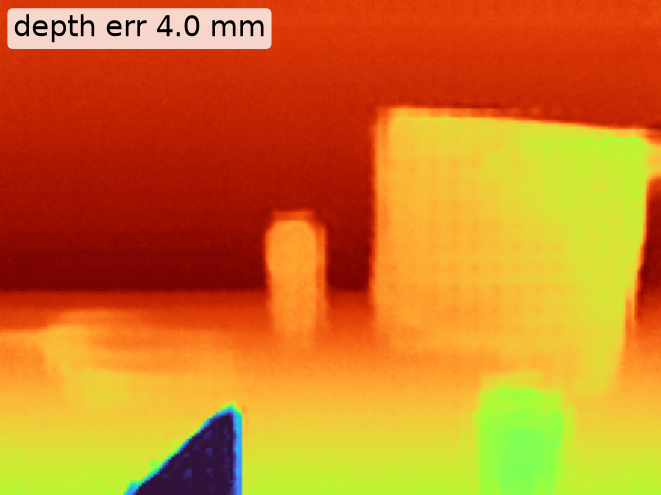} & \pimg{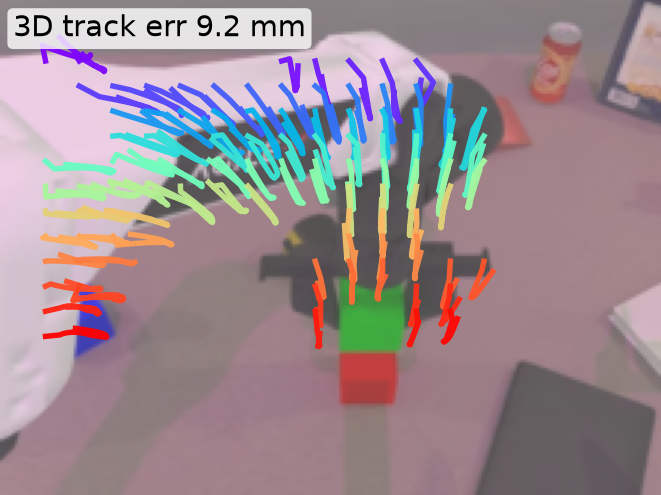} & \pimg{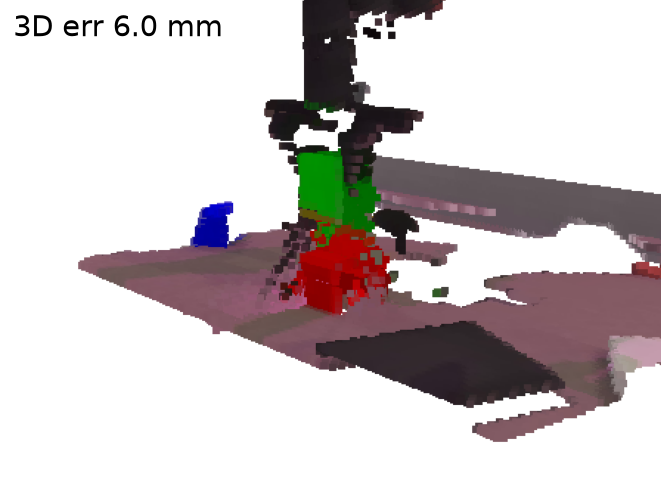} & \pimg{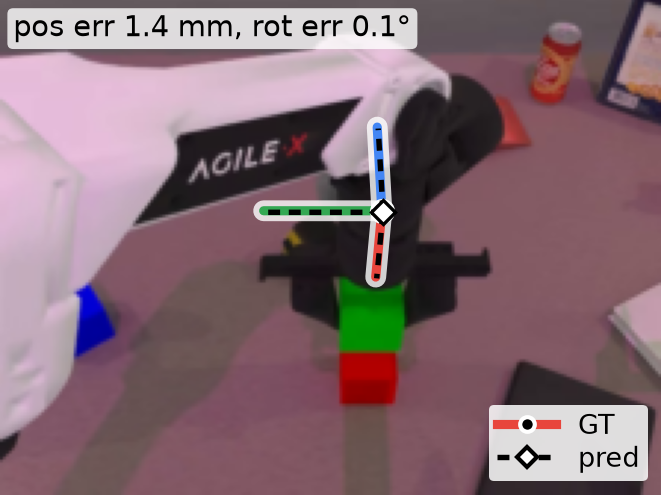} \\
\plabel{Frozen $\pi_{0.5}$} &
\pimg{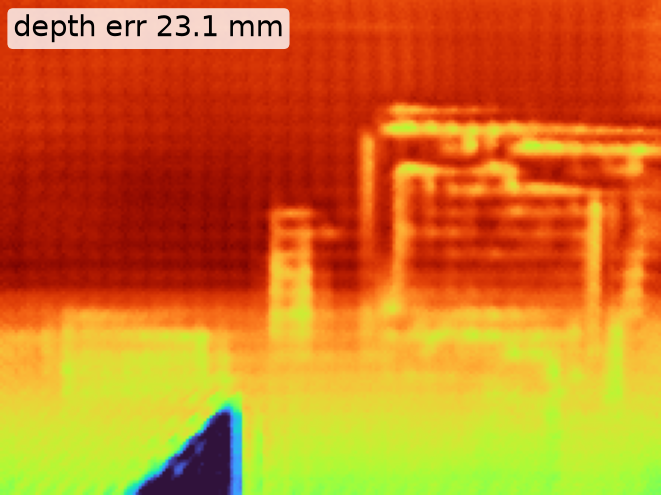} & \pimg{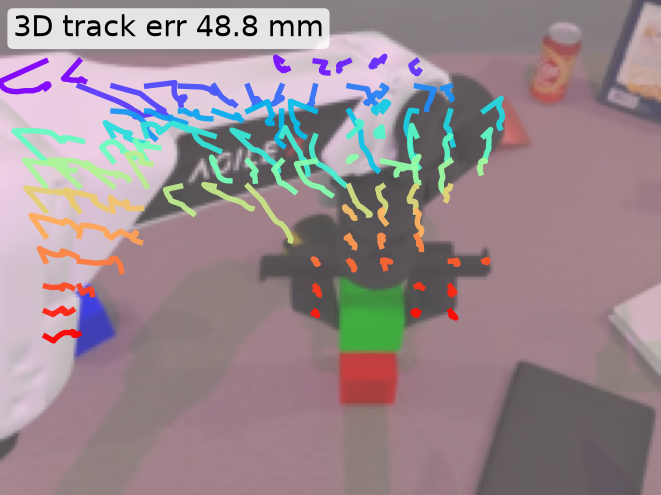} & \pimg{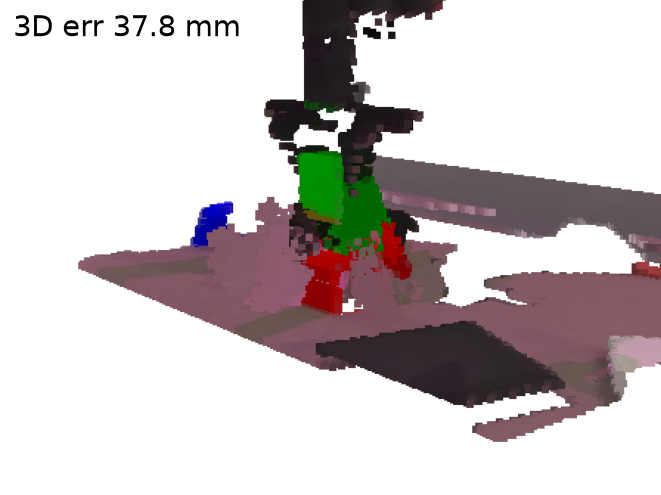} & \pimg{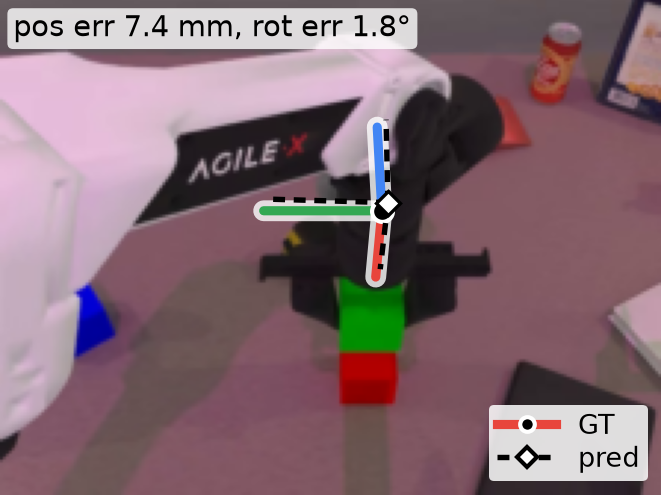} \\
\plabel{Ground Truth} &
\pimg{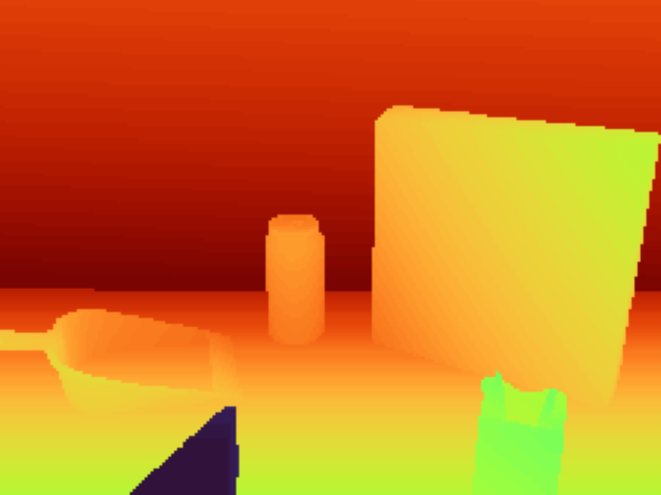} & \pimg{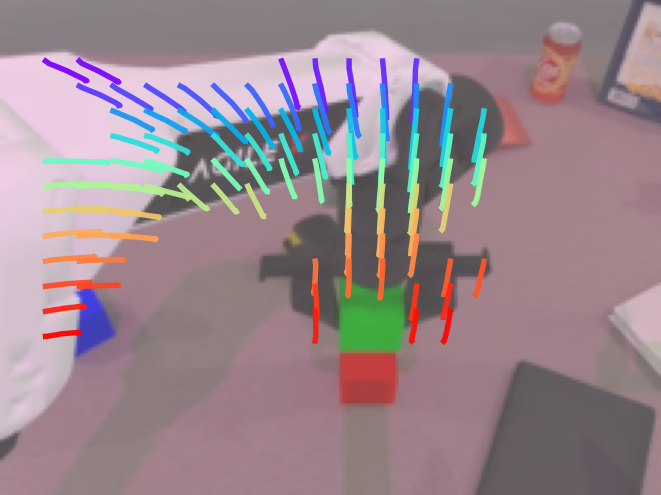} & \pimg{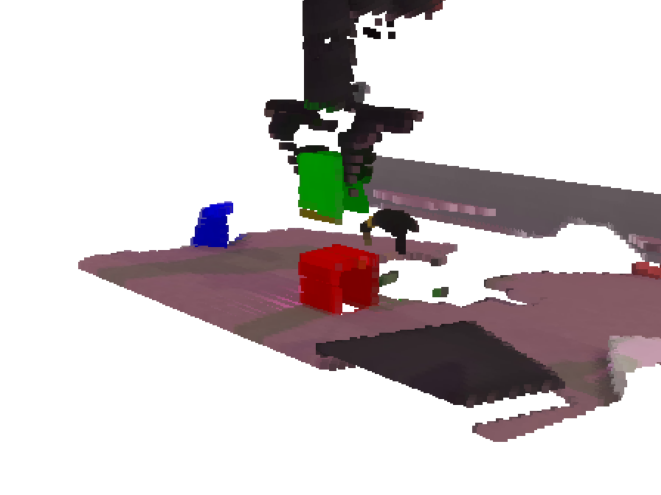} & \pimg{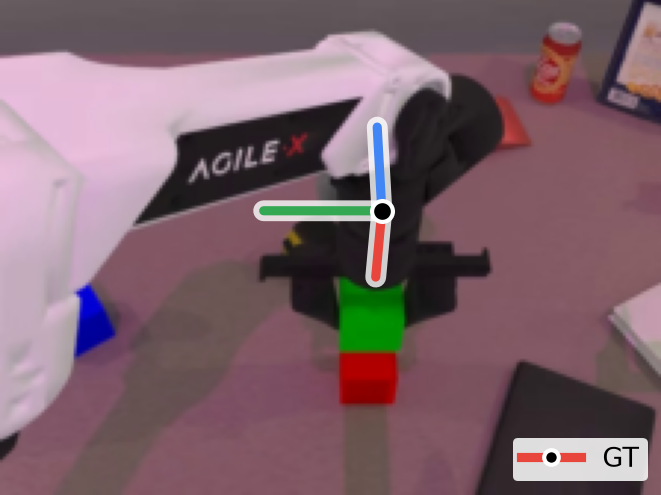} \\
\end{tabular}

\caption{Probing in unseen test environments. The backbone runs once per frame, and the decoder answers non-overlapping queries tiling the image; corner labels give median errors. \textbf{Ours}: decoder trained jointly with the backbone. \textbf{Frozen $\pi_{0.5}$}: the same decoder trained on a frozen $\pi_{0.5}$ backbone. \emph{Depth}: ground truth is quantized at ${\approx}4.7$\,mm; blocking comes from patch-wise decoding. \emph{Motion}: future 3D tracks projected onto the image. \emph{Multi-view}: left-wrist pixels predicted in the head-camera frame, rendered from a virtual viewpoint over the ground-truth head-camera point cloud; errors include points occluded in that view. \emph{Hand-eye}: head-camera pixels predicted in the gripper frame are rigidly aligned with the predicted depth, without ground truth, to recover the gripper pose (8\,cm axes; dashed: prediction, solid: ground truth); the reported error is that of this pose.}
\label{fig:probing}
\end{figure*}

\subsection{RQ5: Does \method make the VLA representation geometry-aware?}
\label{sec:rq5}

\paragraph{The query decoder recovers all four query types in unseen environments.}
The decoder of \method($\pi_{0.5}$) matches the ground truth for all four query types with millimeter-level errors (Fig.~\ref{fig:probing}); motion queries err most, as their targets are only partly determined by the current frame (Sec.~\ref{sec:rq4}).

\paragraph{Query supervision adds geometry beyond the baseline.}
To isolate the effect of query supervision, we add a control run alongside our full model, in which the policy ($\pi_{0.5}$) is frozen and only the \method query decoder is trained. Results show that the decoder performs far worse on all four query types with backbone frozen: action supervision alone does not make this geometry decodable, whereas query supervision does. Together with the gains concentrated on geometry-aware tasks (Sec.~\ref{sec:rq1}), this suggests that \method improves manipulation by making the geometry these tasks depend on directly available in the representation.

\section{Conclusion}
\label{sec:conclusion}


We presented \method, a lightweight framework that injects holistic 4D understanding into robot policies through unified sparse queries. A shared decoder supervises depth, point motion, multi-view correspondence, and hand-eye relations from intermediate VLA representations. A small query budget avoids dense prediction, while removing the decoder after training incurs zero inference overhead. Across RoboTwin~2.0, LIBERO-Plus, and five real-world tasks, \method improves both $\pi_0$ and $\pi_{0.5}$, particularly in geometry-sensitive settings, outperforming WAMs at substantially lower cost. Ablations confirm the benefits of complementary queries, while probing reveals recoverable 4D structure within the policy. These findings show that rich spatiotemporal understanding can emerge from sparse supervision without costly world prediction.

\clearpage


{
\small
\bibliographystyle{ieeenat_fullname}
\bibliography{style/iclr2027_conference}
}

\clearpage
\typeout{CVPR-SUPPLEMENT-START:\thepage}
\setcounter{page}{1}
\renewcommand{\thepage}{S\arabic{page}}
\appendix
\maketitlesupplementary
\section{Implementation details of \method}
\label{app:impl}

This section details the query construction and sampling,
target generation, decoder architecture, and training
configuration of \method.
To ensure fair comparisons, we first optimize the
underlying VLA baselines and then tune \method on top
of these configurations.
All tunable hyperparameters are carefully examined,
with the final settings selected based on empirical
performance and training efficiency.

\subsection{Query construction and sampling}
\label{app:impl:sampling}

\paragraph{Query tokens.}
A query token is constructed from 5 summed parts. The RGB patch is cut from the $224 \times 224$ input
image of the source camera after the backbone's image
preprocessing, with the best choice $p = 14$, which equals the patch size of the SigLIP image encoder in PaliGemma.
The positional encoding applies random Fourier
features~\citep{tancik2020fourierfeaturesletnetworks}
to each pixel coordinate, normalized to $[-1, 1]$.
The learned embedding tables have three entries for
the source camera, $H = 50$ entries for the time
offset, and five entries for the reference frame.
The five terms are summed and layer-normalized.

\paragraph{Query allocation and sampling.}
We sample $M = 96$ queries per frame, evenly allocated across the four query types (24 each).
For spatial sampling, we favor depth discontinuities using an importance map computed from the depth gradient magnitude, smoothed with a $5 \times 5$ box filter, averaged over $4 \times 4$ pixel cells, normalized by its 99th percentile, and clipped to $[0, 1]$.
A cell $i$ with importance $w_i$ is sampled with probability
$\alpha / |\Omega| + (1-\alpha) w_i^\gamma / \sum_{j \in \Omega} w_j^\gamma$,
where $\Omega$ contains valid cells, $\alpha = 0.1$, and $\gamma = 2$.
Pixels are then sampled uniformly within the selected cells.
For temporal sampling, we draw the time offset $\Delta$ uniformly from valid future frames within the 50-frame horizon.

\subsection{Query targets}
\label{app:impl:targets}

\paragraph{Spatial query targets.}
The simulator provides depth maps for all cameras,
camera intrinsics, per-frame camera extrinsics
, and gripper poses.
Targets are computed on the $224 \times 224$ canvas
seen by the backbone, onto which images are resized
with padding.
Depth maps are resized with nearest-neighbor
interpolation, which avoids interpolating depth
across object boundaries.
We set $s = 1.2$\,m and clip normalized coordinates
to $[-1, 1]$.
For the real-world setting, camera parameters and
gripper poses are obtained through hand-eye
calibration and forward kinematics. Depth is obtained through MoGe-3~\citep{kong2026finedetailmonoculargeometryestimation}.

\paragraph{Temporal query targets.}
We generate point tracks offline using
CoTracker3~\citep{karaev24cotracker3} on each camera's
video, with 50-frame windows initialized on a
$20 \times 20$ grid every 25 frames.
For each frame $t$, we use the most recently
initialized window containing $t$ to minimize
accumulated tracking drift.
A motion query selects a point visible at time $t$
using the importance weights above, and samples
$\Delta \in \{1, \ldots, 49\}$ uniformly among
offsets where the point remains visible and its
patch lies within the image.

\begin{table}[t]
\centering
\small
\caption{\textbf{Hyperparameters of \method.}}
\label{tab:app:hparams}
\setlength{\tabcolsep}{4pt}
\begin{tabular*}{\columnwidth}{@{\extracolsep{\fill}}ll@{}}
\toprule
\multicolumn{2}{@{}l}{\emph{Queries}} \\
Queries per frame $M$ & 96 (24 per type) \\
Patch size $p$ & 14 \\
Time offsets $\Delta$ & $0, \ldots, 49$ \\
\midrule
\multicolumn{2}{@{}l}{\emph{Decoder}} \\
Blocks / width & 3 / 1024 \\
Attention heads / dim. & 8 / 128 \\
MLP hidden dim. & 4096 \\
Parameters & 53.7M \\
\midrule
\multicolumn{2}{@{}l}{\emph{Optimization}} \\
Loss weight $\lambda$ & 0.1 \\
Batch size & 64 \\
Optimizer & AdamW ($\beta_1{=}0.9$, $\beta_2{=}0.95$) \\
LR (peak / final) & $1.25{\times}10^{-5}$ / $1.25{\times}10^{-6}$ \\
Schedule & 4k warmup, cosine decay \\
Gradient clipping & 1.0 \\
EMA decay & 0.99 \\
\bottomrule
\end{tabular*}
\end{table}

\subsection{Query decoder}
\label{app:impl:decoder}

The prefix tokens are linearly projected from the
backbone width of $2048$ to the decoder width of $1024$.
Each of the three decoder blocks applies pre-normalized
multi-head cross-attention and a pre-normalized gated
MLP (SwiGLU~\citep{glu}), each with a residual
connection; padding tokens in the prefix are masked
out.
A final layer normalization and a linear layer
produce the $14 \times 14 \times 3$ outputs
of each query.
The decoder has 53.7M parameters.

\subsection{Training details}
\label{app:impl:training}

We initialize the policy from the $\pi_{0.5}$
base checkpoint and the decoder randomly, and train
all parameters jointly with the hyperparameters
in Tab.~\ref{tab:app:hparams}.
Color augmentation is applied to all cameras, following $\pi_{0.5}$ and $\pi_{0}$'s original setting.
For $\pi_0$, we follow the same training protocol
and hyperparameter configuration, replacing only
the base checkpoint.


\begin{table}[t]
\centering
\footnotesize
\caption{Per-task success rates (\%) on all 50 RoboTwin tasks with the $\pi_0$ backbone.
Ours is trained in a single run for 200k steps ($\sim$2.5 epochs) and evaluated with
30 trials per task. The better result in each
setting is shown in \textbf{bold}.}
\label{tab:robotwin_50_pi0}
\setlength{\tabcolsep}{2.5pt}
\renewcommand{\arraystretch}{0.95}
\resizebox{\columnwidth}{!}{%
\begin{tabular}{@{}lrrrr@{}}
\toprule
\multirow{2}{*}{\textbf{Task}} & \multicolumn{2}{c}{$\pi_0$} & \multicolumn{2}{c}{\textbf{\method ($\pi_0$)}} \\
\cmidrule(lr){2-3} \cmidrule(l){4-5}
& Clean & Random & Clean & Random \\
\midrule
\texttt{adjust\_bottle}              & 70.0 & 56.7 & \textbf{100.0} & \textbf{100.0} \\
\texttt{beat\_block\_hammer}         & 80.0 & 73.3 & 80.0 & \textbf{83.3} \\
\texttt{blocks\_ranking\_rgb}        & 80.0 & 83.3 & \textbf{90.0} & \textbf{90.0} \\
\texttt{blocks\_ranking\_size}       & 43.3 & 60.0 & \textbf{63.3} & \textbf{80.0} \\
\texttt{click\_alarmclock}           & \textbf{83.3} & 73.3 & 66.7 & \textbf{80.0} \\
\texttt{click\_bell}                 & \textbf{66.7} & \textbf{83.3} & 40.0 & 63.3 \\
\texttt{dump\_bin\_bigbin}           & \textbf{100.0} & 90.0 & 90.0 & \textbf{93.3} \\
\texttt{grab\_roller}                & 93.3 & 96.7 & \textbf{100.0} & 96.7 \\
\texttt{handover\_block}             & 33.3 & 40.0 & \textbf{80.0} & \textbf{76.7} \\
\texttt{handover\_mic}               & 90.0 & \textbf{100.0} & \textbf{96.7} & 96.7 \\
\texttt{hanging\_mug}                & 20.0 & 20.0 & \textbf{50.0} & \textbf{63.3} \\
\texttt{lift\_pot}                   & 63.3 & 73.3 & \textbf{100.0} & \textbf{93.3} \\
\texttt{move\_can\_pot}              & 56.7 & 53.3 & \textbf{93.3} & \textbf{76.7} \\
\texttt{move\_pillbottle\_pad}       & 73.3 & 66.7 & \textbf{86.7} & \textbf{96.7} \\
\texttt{move\_playingcard\_away}     & 80.0 & 83.3 & \textbf{93.3} & \textbf{93.3} \\
\texttt{move\_stapler\_pad}          & 36.7 & 50.0 & \textbf{83.3} & \textbf{70.0} \\
\texttt{open\_laptop}                & 73.3 & 70.0 & \textbf{80.0} & \textbf{93.3} \\
\texttt{open\_microwave}             & \textbf{70.0} & \textbf{83.3} & 43.3 & 76.7 \\
\texttt{pick\_diverse\_bottles}      & 40.0 & 53.3 & \textbf{60.0} & \textbf{66.7} \\
\texttt{pick\_dual\_bottles}         & 46.7 & 50.0 & \textbf{93.3} & \textbf{83.3} \\
\texttt{place\_a2b\_left}            & 60.0 & 43.3 & \textbf{83.3} & \textbf{83.3} \\
\texttt{place\_a2b\_right}           & 30.0 & 46.7 & \textbf{73.3} & \textbf{70.0} \\
\texttt{place\_bread\_basket}        & 76.7 & 46.7 & \textbf{90.0} & \textbf{86.7} \\
\texttt{place\_bread\_skillet}       & 63.3 & 60.0 & \textbf{90.0} & \textbf{80.0} \\
\texttt{place\_burger\_fries}        & 93.3 & 93.3 & \textbf{100.0} & \textbf{96.7} \\
\texttt{place\_can\_basket}          & 46.7 & 40.0 & \textbf{80.0} & \textbf{86.7} \\
\texttt{place\_cans\_plasticbox}     & 83.3 & 73.3 & \textbf{96.7} & \textbf{93.3} \\
\texttt{place\_container\_plate}     & 93.3 & 100.0 & \textbf{96.7} & 100.0 \\
\texttt{place\_dual\_shoes}          & 63.3 & 50.0 & \textbf{80.0} & \textbf{76.7} \\
\texttt{place\_empty\_cup}           & 100.0 & 96.7 & 100.0 & \textbf{100.0} \\
\texttt{place\_fan}                  & 56.7 & 73.3 & \textbf{86.7} & \textbf{93.3} \\
\texttt{place\_mouse\_pad}           & 43.3 & 40.0 & \textbf{83.3} & \textbf{60.0} \\
\texttt{place\_object\_basket}       & 60.0 & 73.3 & \textbf{86.7} & \textbf{86.7} \\
\texttt{place\_object\_scale}        & 70.0 & 43.3 & \textbf{80.0} & \textbf{73.3} \\
\texttt{place\_object\_stand}        & 70.0 & 76.7 & \textbf{100.0} & \textbf{100.0} \\
\texttt{place\_phone\_stand}         & 70.0 & 56.7 & \textbf{83.3} & \textbf{86.7} \\
\texttt{place\_shoe}                 & 86.7 & 80.0 & \textbf{100.0} & \textbf{96.7} \\
\texttt{press\_stapler}              & 70.0 & \textbf{80.0} & \textbf{73.3} & 76.7 \\
\texttt{put\_bottles\_dustbin}       & 63.3 & 83.3 & \textbf{66.7} & 83.3 \\
\texttt{put\_object\_cabinet}        & 53.3 & 43.3 & \textbf{93.3} & \textbf{76.7} \\
\texttt{rotate\_qrcode}              & 66.7 & 60.0 & \textbf{93.3} & \textbf{86.7} \\
\texttt{scan\_object}                & 46.7 & 23.3 & \textbf{90.0} & \textbf{76.7} \\
\texttt{shake\_bottle}               & 96.7 & 90.0 & \textbf{100.0} & \textbf{100.0} \\
\texttt{shake\_bottle\_horizontally} & \textbf{100.0} & 90.0 & 96.7 & \textbf{96.7} \\
\texttt{stack\_blocks\_three}        & 50.0 & 63.3 & \textbf{86.7} & \textbf{83.3} \\
\texttt{stack\_blocks\_two}          & 80.0 & 93.3 & \textbf{96.7} & \textbf{100.0} \\
\texttt{stack\_bowls\_three}         & 76.7 & 63.3 & 76.7 & \textbf{80.0} \\
\texttt{stack\_bowls\_two}           & 93.3 & 96.7 & \textbf{96.7} & 96.7 \\
\texttt{stamp\_seal}                 & \textbf{60.0} & 53.3 & 53.3 & \textbf{86.7} \\
\texttt{turn\_switch}                & 26.7 & 26.7 & \textbf{43.3} & \textbf{63.3} \\
\midrule
\textbf{Average} & 67.0 & 66.4 & \textbf{83.3} & \textbf{85.0} \\
\bottomrule
\end{tabular}%
}
\end{table}

\section{Additional simulation results}
\label{app:sim_exp_further_analysis}

\begin{table}[t]
\centering
\footnotesize
\caption{Per-task success rates (\%) on all 50 RoboTwin tasks with the $\pi_{0.5}$ backbone.
Ours is trained in a single run for 200k steps ($\sim$2.5 epochs) and evaluated with
30 trials per task. The better result in each
setting is shown in \textbf{bold}.}
\label{tab:robotwin_50_pi05}
\setlength{\tabcolsep}{2.5pt}
\renewcommand{\arraystretch}{0.975}
\resizebox{\columnwidth}{!}{%
\begin{tabular}{@{}lrrrr@{}}
\toprule
\multirow{2}{*}{\textbf{Task}} & \multicolumn{2}{c}{$\pi_{0.5}$} & \multicolumn{2}{c}{\textbf{\method ($\pi_{0.5}$)}} \\
\cmidrule(lr){2-3} \cmidrule(l){4-5}
& Clean & Random & Clean & Random \\
\midrule
\texttt{adjust\_bottle}              & 63.3 & 70.0 & \textbf{100.0} & \textbf{96.7} \\
\texttt{beat\_block\_hammer}         & 76.7 & 93.3 & \textbf{83.3} & 93.3 \\
\texttt{blocks\_ranking\_rgb}        & 86.7 & 86.7 & \textbf{93.3} & \textbf{90.0} \\
\texttt{blocks\_ranking\_size}       & 76.7 & 26.7 & \textbf{80.0} & \textbf{76.7} \\
\texttt{click\_alarmclock}           & 93.3 & 90.0 & \textbf{96.7} & \textbf{100.0} \\
\texttt{click\_bell}                 & 96.7 & 66.7 & \textbf{100.0} & \textbf{96.7} \\
\texttt{dump\_bin\_bigbin}           & 90.0 & 96.7 & \textbf{96.7} & 96.7 \\
\texttt{grab\_roller}                & \textbf{100.0} & \textbf{100.0} & 96.7 & 96.7 \\
\texttt{handover\_block}             & 83.3 & 56.7 & \textbf{90.0} & \textbf{76.7} \\
\texttt{handover\_mic}               & 93.3 & 96.7 & \textbf{100.0} & 96.7 \\
\texttt{hanging\_mug}                & 23.3 & 16.7 & \textbf{46.7} & \textbf{60.0} \\
\texttt{lift\_pot}                   & 100.0 & 86.7 & 100.0 & \textbf{100.0} \\
\texttt{move\_can\_pot}              & 73.3 & 53.3 & \textbf{83.3} & \textbf{83.3} \\
\texttt{move\_pillbottle\_pad}       & 86.7 & 60.0 & \textbf{100.0} & \textbf{93.3} \\
\texttt{move\_playingcard\_away}     & 96.7 & 83.3 & \textbf{100.0} & \textbf{100.0} \\
\texttt{move\_stapler\_pad}          & 60.0 & 43.3 & \textbf{63.3} & \textbf{63.3} \\
\texttt{open\_laptop}                & 86.7 & \textbf{96.7} & \textbf{96.7} & 90.0 \\
\texttt{open\_microwave}             & \textbf{70.0} & \textbf{76.7} & 53.3 & 33.3 \\
\texttt{pick\_diverse\_bottles}      & 20.0 & 70.0 & \textbf{83.3} & \textbf{83.3} \\
\texttt{pick\_dual\_bottles}         & 20.0 & 63.3 & \textbf{96.7} & \textbf{100.0} \\
\texttt{place\_a2b\_left}            & 83.3 & 83.3 & \textbf{86.7} & \textbf{100.0} \\
\texttt{place\_a2b\_right}           & 83.3 & 83.3 & \textbf{93.3} & \textbf{90.0} \\
\texttt{place\_bread\_basket}        & 83.3 & 63.3 & \textbf{93.3} & \textbf{86.7} \\
\texttt{place\_bread\_skillet}       & 86.7 & 66.7 & \textbf{93.3} & \textbf{83.3} \\
\texttt{place\_burger\_fries}        & 96.7 & 86.7 & 96.7 & \textbf{100.0} \\
\texttt{place\_can\_basket}          & 53.3 & 63.3 & \textbf{80.0} & \textbf{80.0} \\
\texttt{place\_cans\_plasticbox}     & 90.0 & 83.3 & \textbf{100.0} & \textbf{100.0} \\
\texttt{place\_container\_plate}     & 90.0 & 93.3 & \textbf{100.0} & \textbf{96.7} \\
\texttt{place\_dual\_shoes}          & \textbf{93.3} & 73.3 & 73.3 & 73.3 \\
\texttt{place\_empty\_cup}           & 96.7 & 100.0 & 96.7 & 100.0 \\
\texttt{place\_fan}                  & 86.7 & 86.7 & \textbf{100.0} & \textbf{93.3} \\
\texttt{place\_mouse\_pad}           & 63.3 & 40.0 & \textbf{73.3} & \textbf{76.7} \\
\texttt{place\_object\_basket}       & 60.0 & 76.7 & \textbf{86.7} & \textbf{83.3} \\
\texttt{place\_object\_scale}        & 83.3 & 80.0 & \textbf{93.3} & \textbf{93.3} \\
\texttt{place\_object\_stand}        & 90.0 & 86.7 & \textbf{96.7} & \textbf{96.7} \\
\texttt{place\_phone\_stand}         & 83.3 & 80.0 & \textbf{90.0} & \textbf{96.7} \\
\texttt{place\_shoe}                 & 96.7 & 93.3 & \textbf{100.0} & \textbf{100.0} \\
\texttt{press\_stapler}              & 80.0 & 83.3 & \textbf{90.0} & \textbf{93.3} \\
\texttt{put\_bottles\_dustbin}       & 76.7 & 80.0 & \textbf{100.0} & \textbf{93.3} \\
\texttt{put\_object\_cabinet}        & 83.3 & 80.0 & \textbf{86.7} & \textbf{90.0} \\
\texttt{rotate\_qrcode}              & 80.0 & 86.7 & \textbf{96.7} & \textbf{90.0} \\
\texttt{scan\_object}                & 80.0 & 66.7 & \textbf{93.3} & \textbf{90.0} \\
\texttt{shake\_bottle}               & 96.7 & 96.7 & \textbf{100.0} & \textbf{100.0} \\
\texttt{shake\_bottle\_horizontally} & 100.0 & 100.0 & 100.0 & 100.0 \\
\texttt{stack\_blocks\_three}        & 83.3 & 76.7 & \textbf{93.3} & \textbf{86.7} \\
\texttt{stack\_blocks\_two}          & \textbf{100.0} & 100.0 & 96.7 & 100.0 \\
\texttt{stack\_bowls\_three}         & 76.7 & 70.0 & \textbf{93.3} & \textbf{76.7} \\
\texttt{stack\_bowls\_two}           & 93.3 & \textbf{96.7} & \textbf{100.0} & 93.3 \\
\texttt{stamp\_seal}                 & 73.3 & 53.3 & \textbf{86.7} & \textbf{100.0} \\
\texttt{turn\_switch}                & 56.7 & 53.3 & \textbf{83.3} & \textbf{66.7} \\
\midrule
\textbf{Average} & 79.9 & 76.3 & \textbf{90.7} & \textbf{89.1} \\
\bottomrule
\end{tabular}%
}
\end{table}

As shown in Tab.~\ref{tab:robotwin_50_pi0} and Tab.~\ref{tab:robotwin_50_pi05}, we do another full run on RoboTwin 2.0~\citep{chen2025robotwin}, with 200k steps and batch size 64.

\section{Real-world experimental setup}
\label{app:real_world_setup}

\begin{figure*}[t]
    \centering
    \includegraphics[width=\linewidth]{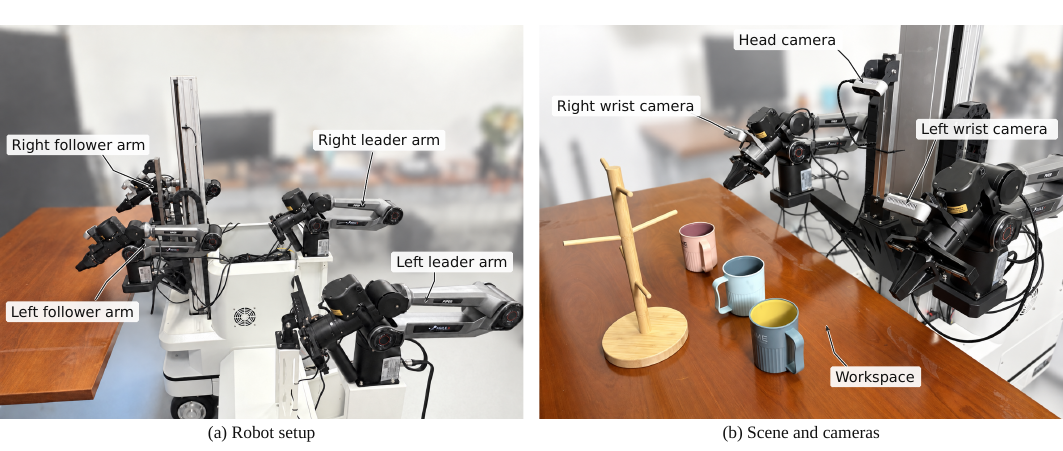}
    \caption{Dual-arm PiPER-X mobile manipulator platform.}
    \label{fig:real_world_platform}
\end{figure*}

This section describes the physical platform (App.~\ref{app:hardware}), demonstration data collection procedure (App.~\ref{app:real_data_collection}), evaluation protocols (App.~\ref{app:real_evaluation}), and task-specific settings (App.~\ref{app:real_world_tasks}) used in all real-world experiments.

\subsection{Robot platform}
\label{app:hardware}

All real-world experiments are conducted on a dual-arm PiPER-X mobile manipulator platform, as shown in Fig.~\ref{fig:real_world_platform}. Each arm consists of six joints and is equipped with a parallel-jaw gripper controlled by a continuous-valued gripper command. Three synchronized Intel RealSense D435 cameras---one head-mounted camera and one wrist-mounted camera on each arm---are mounted on the platform.

\begin{figure*}[t]
    \centering
    \includegraphics[width=0.85\textwidth]{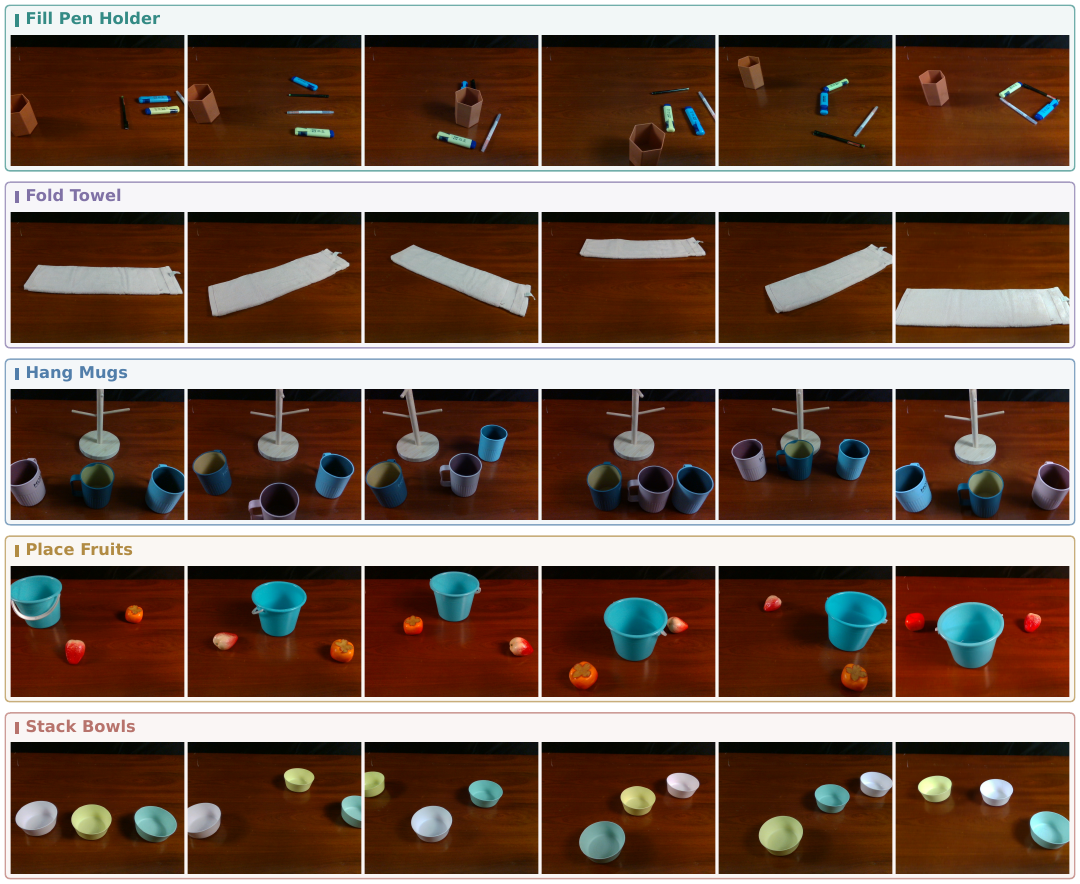}
    \caption{\textbf{Diverse initial layouts for real-world evaluation.} Each row shows six example evaluation layouts for one task, illustrating varied object positions, orientations, and relative arrangements within the head camera's view.}
    \label{fig:real_world_eval_layouts}
\end{figure*}

\subsection{Data collection}
\label{app:real_data_collection}

Our data collection and deployment pipelines are built upon LeRobot~\citep{cadene2024lerobot}. For each task, we collect 50 successful teleoperated demonstration trajectories. Each trajectory contains synchronized multi-view RGB observations with size $640{\times}480$, robot proprioception and joint-space action streams at $30\,\mathrm{Hz}$. When required by a policy's action representation, we convert joint-space actions to end-effector-space actions using forward kinematics.

\begin{figure*}[t]
    \centering
    \includegraphics[width=0.82\linewidth]{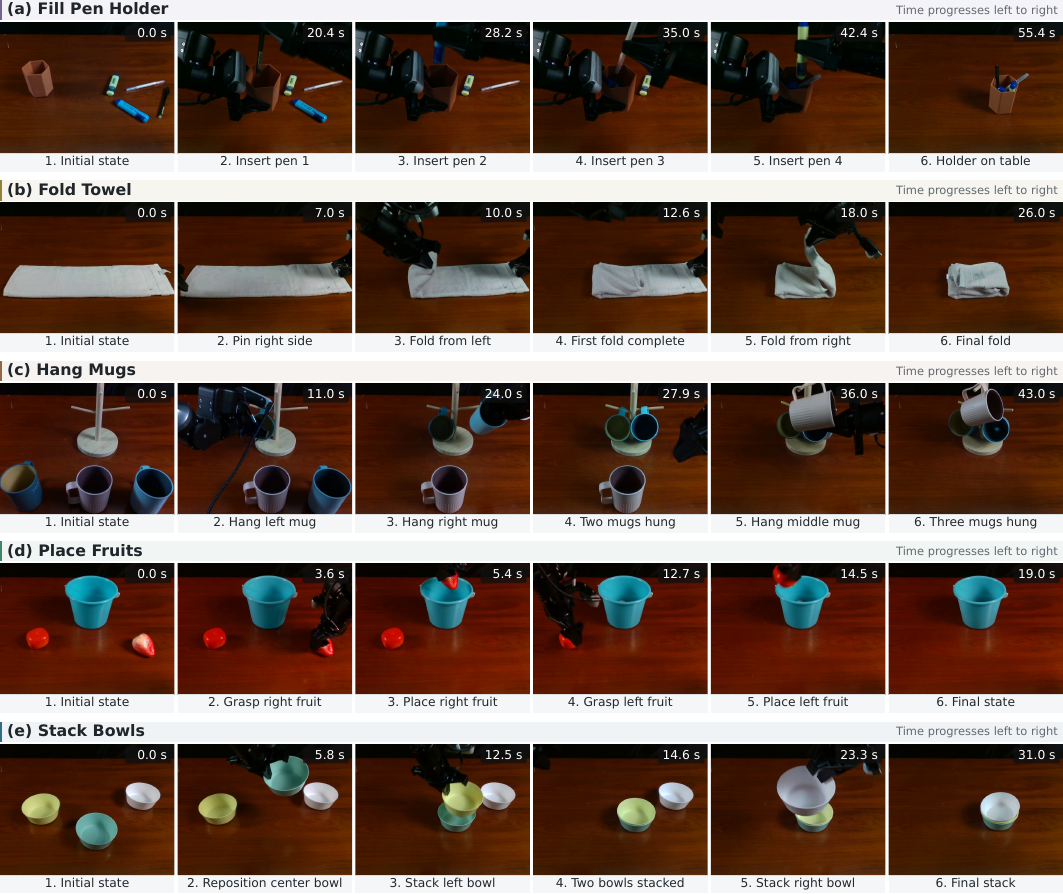}
    \caption{The five real-world tasks. For each task we show representative keyframes from a successful rollout.}
    \label{fig:real_world_tasks}
\end{figure*}

\begin{table*}[t]
\centering
\small
\caption{Per-trial progress-scoring rubrics for real-world tasks (0--100 points).}
\label{tab:combined_score_rubrics}
\begin{tblr}{
  colspec = {Q[c,m,2.5cm] Q[c,m,1.2cm] X[l,m]},
  hline{1,2,8,11,15,18,Z},
  rowsep = 0pt,
  vspan = even,
}
Task & Score & Condition \\
\SetCell[r=6]{c} Fill Pen Holder
 & 0   & No pen is correctly inserted into the pen holder. \\
 & 10  & One pen is correctly inserted tip-up into the upright pen holder. \\
 & 25  & Two pens are correctly inserted into the pen holder. \\
 & 40  & Three pens are correctly inserted into the pen holder. \\
 & 90  & All four pens are inserted into the upright pen holder, but the robot has not returned home. \\
 & 100 & All four pens are inserted, the pen holder is upright, and the robot has returned home. \\
\SetCell[r=3]{c} Fold Towel
 & 0   & The towel does not form a rectangular shape, or the two-step folding procedure is not followed. \\
 & 50  & The two-step folding procedure is followed and the towel is roughly rectangular, but one or more sides are visibly misaligned. \\
 & 100 & The two-step folding procedure is followed and the towel forms a clean square or rectangular fold with the sides properly aligned. \\
\SetCell[r=4]{c} Hang Mugs
 & 0   & No mug is hung on the mug rack. \\
 & 20  & One mug is hung on the mug rack. \\
 & 50  & Two mugs are hung on the mug rack. \\
 & 100 & Three mugs are hung on the mug rack. \\
\SetCell[r=3]{c} Place Fruits
 & 0   & No fruit is placed into the bucket. \\
 & 30  & One fruit is placed into the bucket. \\
 & 100 & Both fruits are placed into the bucket. \\
\SetCell[r=3]{c} Stack Bowls
 & 0   & Fewer than two bowls are stacked in the center. \\
 & 30  & Two bowls are stacked in the center. \\
 & 100 & All three bowls are stacked in the center. \\
\end{tblr}
\end{table*}

\subsection{Evaluation protocols}
\label{app:real_evaluation}

To control variation in scene initialization and enable paired comparison across methods, we construct a fixed set of 25 evaluation configurations for each task and reuse the same configurations for every evaluated method. Before every rollout, the evaluator manually restores the objects to the corresponding reference configuration using a visual layout overlay, following the layout-replay protocol of RoboDojo~\citep{chen2026robodojo}.

The evaluation configurations span diverse object positions, orientations, and relative arrangements within the tabletop workspace visible to the head camera. These variations require policies to adapt their grasps and manipulation motions to different initial layouts, including changes in towel orientation, container placement, and the spatial arrangements of mugs and bowls. Fig.~\ref{fig:real_world_eval_layouts} shows representative configurations for all five tasks.

We report success rate (SR) and a task-specific progress score (PS). Each trial is assigned a score in $[0,100]$ according to the task-specific rubrics in Tab.~\ref{tab:combined_score_rubrics}. A trial is successful if and only if it receives 100 points. SR is the proportion of successful trials, and PS is the mean per-trial score across the 25 evaluation configurations. Both metrics are computed under the same fixed set of initial configurations for all compared methods.

All methods are deployed using synchronous chunked inference. At each policy query, the model receives the latest observation and predicts an action chunk, which is executed open-loop in its entirety. The next policy query then uses the latest available observation. We use an action horizon of $H=50$ for the $\pi$-series VLA policies and $H=32$ for WAM-based policies, at the control frequency of $30\,\mathrm{Hz}$.

\subsection{Real-world task suite}
\label{app:real_world_tasks}

We use five real-world manipulation tasks, spanning complementary sources of difficulty: rigid-object pick-and-place, repeated rigid-object manipulation under a stacking constraint, long-horizon bimanual insertion, deformable-object folding, and precision bimanual hanging, as shown in Fig.~\ref{fig:real_world_tasks}. Three of the five tasks (Stack Bowls, Fill Pen Holder, and Hang Mugs) follow the corresponding task definitions from the RoboDojo real-world benchmark \citep{chen2026robodojo}; the remaining two (Place Fruits and Fold Towel) are defined in-house following the same documentation format. For each task, we specify the language prompt, a brief task description, and a graded scoring rubric listed in Tab.~\ref{tab:combined_score_rubrics}.

\paragraph{Fill Pen Holder} A pen holder and four pens, consisting of two flat highlighters and two slim gel pens, are placed on the table. The robot must pick up the holder and insert all four pens into it. The holder is initialized in the left part of the head camera's field of view, while the pens are initialized in the right part with randomized positions and orientations. The prompt is \textit{Pick up the pen holder and place all the pens into it.}

\paragraph{Fold Towel} A flattened towel is placed on the table, and the robot must perform two sequential folds to form a compact rectangular shape using bimanual manipulation. The required procedure consists of two steps. First, the right arm holds down the right side of the towel while the left arm grasps its lower-left portion and folds it in half. Second, the left arm remains raised while the right arm lifts slightly and opens its gripper, then grasps the middle of the towel's right side and folds it in half again. The prompt is \textit{Fold the towel.}

\paragraph{Hang Mugs} Three mugs and a mug rack are placed on the table, and the robot must pick up the mugs and hang them securely on the rack. The prompt is \textit{Hang the mugs on the mug rack.}

\paragraph{Place Fruits} A strawberry and a persimmon are placed on the table, and the robot must place both fruits into a bucket. The prompt is \textit{Place both fruits into the bucket.}

\paragraph{Stack Bowls} Three bowls are placed on the table, and the robot must manipulate them one at a time to form a single vertical stack at the center of the workspace. The bowls are initialized with randomized left-to-right ordering and positions within the head camera's field of view. The prompt is \textit{Stack the bowls on the table.}

\end{document}